\documentclass[letterpaper, 10 pt, journal, twoside]{IEEEtran} 

\IEEEoverridecommandlockouts                              

\usepackage[dvipsnames]{xcolor}
\definecolor{dark-green}{RGB}{12,80,12}

\usepackage{dsfont}
\usepackage{amsmath}
\usepackage{amssymb}
\usepackage{cuted}
\usepackage{multirow} 
\usepackage{multicol}
\usepackage{tikz}
\usepackage[shortlabels, inline]{enumitem}
\newcommand{\tableref}[1]{Table~\ref{#1}}
\usepackage{tabularx}
\newcolumntype{C}[1]{>{\centering\let\newline\\\arraybackslash\hspace{0pt}}m{#1}} 
\newcolumntype{L}[1]{>{\let\newline\\\arraybackslash\hspace{0pt}}m{#1}} 

\usepackage[mathscr]{euscript}

\usepackage[export]{adjustbox}
\newcolumntype{P}[1]{>{\centering\arraybackslash}p{#1}}

\iffalse 
  \newcommand{\todo}[1]{\noindent}
\else
  \newcommand{\todo}[1]{\textcolor{red}{\bf [Todo: #1]}}
\fi

\usepackage{booktabs}
\usepackage{xspace}
\usepackage{caption}
\usepackage{graphicx} 
\usepackage{color}
\usepackage{siunitx}
\usepackage[hang,flushmargin,symbol]{footmisc}
\renewcommand{\thefootnote}{\fnsymbol{footnote}}
\usepackage{microtype}
\usepackage{rotating}
\newcommand{\secref}[1]{Sec.~\ref{#1}}
\renewcommand{\eqref}[1]{Eq.~(\ref{#1})}
\newcommand{\figref}[1]{Fig.~\ref{#1}}
\newcommand{\tabref}[1]{Tab.~\ref{#1}}

\usepackage{tabularx}
\usepackage{colortbl} 
\newcolumntype{Y}{>{\centering\arraybackslash}X}
\newcolumntype{Z}{>{\raggedleft\arraybackslash}X}

\usepackage{array}
\usepackage{multirow}
\usepackage{pifont}
\usepackage{cite}
\usepackage[export]{adjustbox}
\usepackage[resetlabels]{multibib}
\newcites{New}{References}
\usepackage[linesnumbered,ruled,vlined]{algorithm2e}

\newcommand\rebuttal[1]{\textcolor{black}{#1}}
\usepackage{tabu}
\usepackage{float}

\usepackage{pgfplotstable}
\usepackage{tikz}
\usepackage{soul}

\usepackage{url}

\newcommand{\reffig}[1]{Fig.~\ref{#1}}
\newcommand{\refsec}[1]{Sec.~\ref{#1}}
\newcommand{\refalg}[1]{Algorithm~\ref{#1}}

\DeclareSIUnit{\rad}{rad}

\renewcommand{\baselinestretch}{0.985}

\title{\LARGE \bf
3D Multi-Object Tracking Using Graph Neural Networks with Cross-Edge Modality Attention
}

\author{Martin B\"uchner\thanks{Martin B\"uchner and Abhinav Valada are with the Robot Learning Lab, Department of Computer Science, University of Freiburg, Germany.}, 
and Abhinav Valada\footnotemark[1] 
\thanks{This work was funded by the Eva Mayr-Stihl Foundation.}}

\author{
Martin B\"uchner 
and Abhinav Valada
\thanks{Department of Computer Science, University of Freiburg, Germany.}%
\thanks{This work was funded by the European Union’s Horizon 2020 research and innovation program under grant agreement No 871449-OpenDR.}
\thanks{This paper provides supplementary material at
\textit{https://arxiv.org/abs/2203.10926}.}
}

\begin{document}

\maketitle
\thispagestyle{empty}
\pagestyle{empty}

\begin{abstract}
Online 3D multi-object tracking (MOT) has witnessed significant research interest in recent years, largely driven by demand from the autonomous systems community. However, 3D offline MOT is relatively less explored. Labeling 3D trajectory scene data at a large scale while not relying on high-cost human experts is still an open research question. In this work, we propose Batch3DMOT which follows the tracking-by-detection paradigm and represents real-world scenes as directed, acyclic, and category-disjoint tracking graphs that are attributed using various modalities such as camera, LiDAR, and radar. We present a multi-modal graph neural network that uses a cross-edge attention mechanism mitigating modality intermittence, which translates into sparsity in the graph domain. Additionally, we present attention-weighted convolutions over frame-wise \mbox{k-NN} neighborhoods as suitable means to allow information exchange across disconnected graph components. \rebuttal{We evaluate our approach using various sensor modalities and model configurations on the challenging nuScenes and KITTI datasets. Extensive experiments demonstrate that our proposed approach yields an overall improvement of 3.3\% in the AMOTA score on nuScenes thereby setting the new state-of-the-art for 3D tracking and further enhancing false positive filtering}.
\end{abstract}

\section{Introduction}

3D multi-object tracking (MOT) is an essential component of the scene understanding pipeline of autonomous robots. It aims at inferring associations between occurrences of object instances at different time steps in order to predict plausible 3D trajectories. These trajectories are then used in various downstream tasks such as trajectory prediction~\cite{radwan2020multimodal} and navigation~\cite{mittal2019vision}. Tracking multiple objects under real-time constraints in an online setting is challenging due to both intermediate track prediction when facing false negatives and robust false positive filtering. Owing to recent advances in LiDAR-based object detection~\cite{centerpoint}, the 3D tracking task has also seen significant performance improvements. 

Real-world deployment of these online methods in areas such as autonomous driving poses several challenges. When requiring regulatory approval, its robust behavior must be demonstrated on large sets of reference data which is arduous to obtain due to the lack of extensive ground truth. Therefore, performing high-quality offline labeling of real-world traffic scenes provides the means to test online methods on a larger scale and further sets a benchmark for what is within the realms of possibility for online methods. With respect to generating pseudo ground truth, our proposed method aims at minimizing the number of false positive trajectories at high recalls.

\begin{figure}
    \centering
     \includegraphics[width=1.02\linewidth]{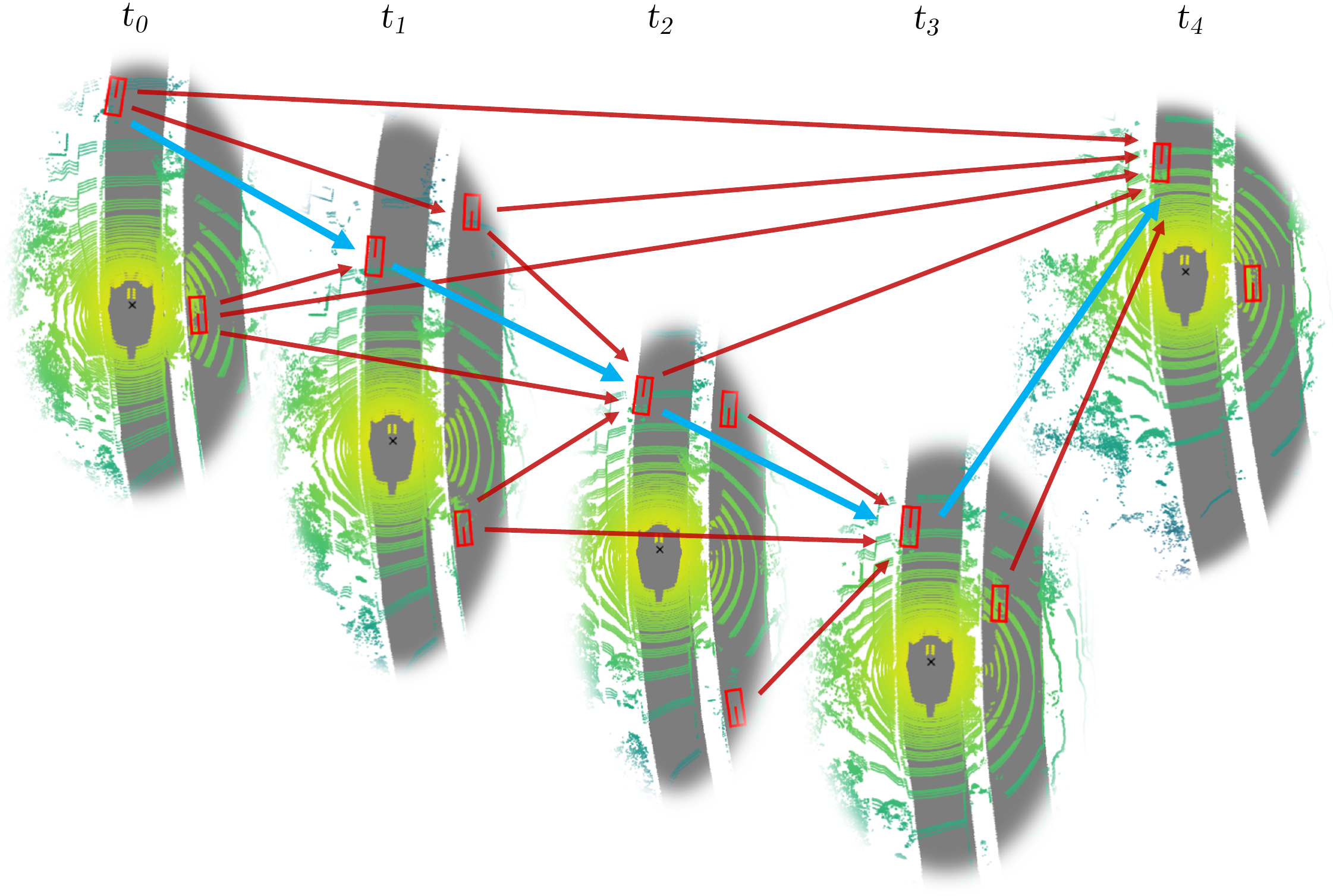}
     \caption{Birds-eye view visualization of a 3D offline tracking scenario showing the road surface and LiDAR point clouds recorded at different time steps. The goal in tracking is to find plausible chains of edges connecting objects across time that best explain the evolution of an object instance. This representation only shows the edges accompanying a single object instance.}
     \label{fig:paper-teaser}
     \vspace{-0.4cm}
\end{figure}

In this paper, we present Batch3DMOT, an offline 3D tracking framework that follows the tracking-by-detection paradigm and utilizes multiple sensor modalities (camera, LiDAR, radar) to solve a multi-frame, multi-object tracking objective. Sets of 3D object detections per frame are first turned into attributed nodes. In order to learn offline 3D tracking, we employ a graph neural network (GNN) that performs time-aware neural message passing with intermediate frame-wise attention-weighted neighborhood convolutions. Different from popular Kalman-based approaches such as AB3DMOT~\cite{ab3dmot}, which essentially tracks objects of different semantic categories independently, our method uses a single model that operates on category-disjoint graph components. As a consequence, it leverages inter-category similarities to improve tracking performance. While Bras\'o~\textit{et~al.} were able to solve a single-category 2D offline tracking objective using graph neural networks~\cite{braso2020learning}, this work focuses on the 3D MOT task while ensuring balanced performance across different semantic categories.

When evaluating typically used modalities such as LiDAR, we can make a striking observation: On the one hand, detection features such as bounding box size or orientation are consistently available across time. A similar observation can be made for camera features, even if the object is (partially) occluded. On the other hand, sensor modalities such as LiDAR or radar do not necessarily share this availability. Due to their inherent sparsity, constructing a feature, e.g., for faraway objects, is typically impractical as it does not serve as a discriminative feature that can be used in tracking. This potential modality intermittence translates to sparsity in the graph domain, which is tackled in this work using our proposed cross-edge modality attention. This enables an edge-wise agreement on whether to include the particular modality in node similarity finding.

\rebuttal{Our main contributions can be summarized as follows:
\begin{itemize}[noitemsep, topsep=0pt]
\item A novel multimodal GNN framework for offline 3D multi-object tracking on multi-category tracking graphs including k-NN neighborhood attention across semantic graph components.
\item A cross-edge attention mechanism that uses intermittent sensor data to substantiate the differentiation between active and inactive edges.
\item Methodology and pre-processing pipeline for constructing category-disjoint tracking graphs over multiple timesteps as well as a novel agglomerative trajectory clustering scheme for effective trajectory generation.
\item Extensive evaluations and ablation study on the nuScenes~\cite{nuscenes} and KITTI~\cite{kitti} datasets using different detection approaches.
\item The code and pre-trained models are publicly available at \url{http://batch3dmot.cs.uni-freiburg.de}.
\end{itemize}}

\section{Related Work}
\label{sec:relatedWork}

Multi-object tracking (MOT) can be categorized into online and offline settings. Whereas online methods are limited to using past and current data, offline methods can efficiently leverage future data to find solutions to the global data association problem. Besides a temporal categorization, MOT can be applied in the 2D~\cite{bytetrack, hurtado2020mopt, valverde2021there} or the 3D domain~\cite{eagermot, ab3dmot, centerpoint, frossard}, exploiting either 2D or 3D object detections. Finally, two commonly followed approaches involve the tracking-by-detection paradigm~\cite{bytetrack, braso2020learning, centerpoint} and joint object detection and tracking~\cite{hurtado2020mopt}. In this section, we briefly review offline 2D MOT and discuss selected 3D MOT methods, relevant to our work.

{\parskip=5pt
\noindent\textit{2D Multi-Object Tracking}:
2D MOT has been extensively studied by the scientific community. Often, both online and offline methods are jointly evaluated on a single benchmark \cite{mot20}. Typically, the underlying datasets comprise largely static scenes with various angles of view and at high frame rates showing a single object category. Most offline methods formulate MOT as a graph association problem solved using optimization techniques from graph and network theory, e.g., min-cost flow optimization~\cite{learning_optimal_parameters_wang}, min-clique graphs~\cite{gmcp_zamir}, lifted multicuts~\cite{lifted_multicut_tang}, and lifted disjoint paths~\cite{hornakova2020lifted}. Exploiting deep learning, several methods investigate either pair-wise appearance similarities~\cite{gmcp_zamir} or specifically focus on learning the data association task via end-to-end backpropagation~\cite{learning_optimal_parameters_wang}. The advent of graph neural networks (GNNs) further allows learning higher-order similarities on graph structures. Along the same line of research, the offline tracker NMPTrack~\cite{braso2020learning} proposes neural message passing to effectively represent both past and future of each detection by leveraging a time-aware prior.
Inspired by this idea, we also exploit future information via a time-aware prior.
}

{\parskip=5pt
\noindent\textit{3D Multi-Object Tracking}:
Compared to popular 2D datasets, available 3D MOT datasets are more challenging since they involve intricate sensor motion and significantly smaller frame rates~\cite{nuscenes, kitti, mot20}. On the other hand, 3D instance detection at varying depth levels allows to effectively resolve occlusion.

Conventional 3D tracking-by-detection approaches mostly rely on bounding box information, following Bewley \textit{et al.}~\cite{sort} in using a Kalman filter as a motion model and the Hungarian algorithm for bipartite data association. While Weng~\textit{et~al.}~\cite{ab3dmot} use 3D-IoU as the matching criterion, Chiu~\textit{et~al.}~\cite{chiu2020probabilistic} employ the Mahalanobis distance, estimate the initial noise and state covariance of the Kalman filter from the training set, and choose a greedy algorithm for data association. Different from these approaches, CenterPoint~\cite{centerpoint} is a 3D object detection model that utilizes a keypoint detector to first predict object centers and then perform regression of object attributes, e.g., dimension, orientation, and velocity. Additionally, CenterPoint proposes 3D tracking based on the closest-point matching of object velocity vectors. Combining both online and offline paradigms, FG-3DMOT~\cite{poschmann2020factor} casts the tracking problem as a factor graph over 3D object detections represented as a Gaussian mixture model in order to find probabilistic trajectory assignments. Other models incorporate 2D object detections as they are less prone to occlusions than their 3D counterparts~\cite{gnn3dmot, eagermot}.

In addition to bounding box information, multiple works include 2D/3D appearance features to substantiate pair-wise affinity representation~\cite{chiu2021probabilistic}. Deep learning allows to learn semantic features via encoding: Popular methods~\cite{frossard, gnn3dmot} use image classification networks as encoders for representing image data and PointNet~\cite{pointnet} architectures for learning point cloud features. Regarding the inclusion of appearance features, GNN3DMOT~\cite{gnn3dmot} is the work most similar to ours. It concatenates encoded modality features before regressing an affinity matrix used for bipartite matching. Recent state-of-the-art approaches~\cite{gnn3dmot, zaech2021learnable} facilitate graph neural networks to capture higher-order artefacts on graph structures. OGR3MOT~\cite{zaech2021learnable} follows NMPTrack~\cite{braso2020learning} in using neural message passing but solves the online 3DMOT problem while leveraging Kalman state predictions for improved track representations. 

Although the aforementioned 3D tracking methods show remarkable results in the online setting, they are insufficient in the scenario of offline 3D MOT due to moderate false positive handling. The only two methods solving 3D offline tracking~\cite{frossard, poschmann2020factor} do not provide publicly available implementations, nonetheless we show a comparison on the KITTI benchmark dataset~\cite{kitti}. Different to these aforementioned offline approaches we utilize a graph neural network to learn the tracking task using higher-order node similarities. 
\rebuttal{Our approach differs from NMPTrack~\cite{braso2020learning} by introducing a novel modality and node representation scheme relevant for 3D tracking and a novel agglomerative trajectory clustering scheme that yields high recall and fewer false positives. Different from OGR3MOT~\cite{zaech2021learnable}, we include multiple sensor modalities and model trajectories based on object similarity instead of exploiting Kalman filters for predictive track representation in online tracking.}}

\section{Technical Approach}
\label{sec:proposedApproach}

Following the tracking-by-detection paradigm, we turn a set of detections per frame $O_t = \{o_1, . . . , o_n\}$ into nodes on a directed acyclic graph $G = (V,E)$ that holds an ordered set of frames. The graph consists of a set of nodes $j \in V$ that are connected via directed edges $E \subseteq \{(j,i) \mid (j,i) \in V^{2} \text{ and } j \neq i\}$, where edges are directed in a forward-time manner. Instead of using detection edges, we follow the approach of Bras\'o~\textit{et~al.}~\cite{braso2020learning} in collapsing them. As a consequence, nodes always reside in a specific frame and edges only connect nodes at different timestamps while satisfying $t_j < t_i$. Tracking multiple objects in the offline setting entails finding a set of edge-disjoint trajectories $\mathcal{T} = \{T_1, \ldots, T_m\}$ that represents the most plausible association of detections over time. Since our approach involves learning on graph-structured data, both nodes and edges are attributed. We refer to the node feature matrix as $\mathbf{X} = [\mathbf{h}_1, ..., \mathbf{h}_N]^{T} \in \mathbb{R}^{N \times D}$ where $h_i \in \mathbb{R}^{D}$ represents a single node feature. Similarly, we denote the edge features $\mathbf{X}_e = [..., \mathbf{h}_{ji},...]^{T} \in \mathbb{R}^{|E| \times D_e}$, where $\mathbf{h}_{ji}$ is the edge feature associated with edge $(j,i)$.


\subsection{Feature Representation}

In typical tracking scenarios such as autonomous driving, we are confronted with a multitude of sensor modalities such as camera, LiDAR, radar, or even thermal images. While the detections are often derived only from a single sensor modality such as LiDAR or camera, the entirety of modalities can still be utilized for improved similarity finding of detections in the tracking task. Our approach fuses 3D pose \& motion features (3D-PM) from bounding boxes with 2D as well as 3D appearance features from (surround) cameras (2D-A), LiDAR (3D-AL) as well as radar sensors (3D-R). Different from tracking in the image plane, 3D bounding box information essentially represents a more discriminative feature in 3D tracking due to available depth information~\cite{gnn3dmot}. Most importantly, this simplifies re-association after false negatives (FN) generated by occlusions or missed detections but also eases the identification of false positives (FP). Instead of solely exploiting bounding box information in terms of relative node differences for an initial edge feature~\cite{braso2020learning}, the 3D-PM feature constitutes the primary node feature in the proposed approach.

\subsubsection{3D Pose and Motion Feature}
We turn each 3D bounding box in the set of detections into an explicit 3D-PM feature without further encoding:
\begin{equation}
\mathbf{h}_{\text{PM,}i} =[x,y,z,w,l,h,\gamma,v_x,v_y,\boldsymbol{c},\mathcal{S},t]^{T} \in \mathbb{R}^{11+C},
\end{equation}
where $x,y,z$ denotes the 3D object center position in ego-vehicle coordinates. The 3D bounding box dimensions are given by $w,l,h$, while the box orientation is expressed by the yaw angle $\gamma$ about the positive z-axis w.r.t. the ego-vehicle frame. Similarly, $v_x,v_y$ describe the relative object center velocity in the x-y-plane. In addition, a one-hot class vector $\mathbf{c}$ over $C$ classes is appended to encode semantic categories. The detection confidence score $\mathcal{S} \in [0,1]$ provides an additional means to differentiate between plausible and implausible detections. Finally, a relative timestamp is included. We choose ego-vehicle coordinates over global map coordinates to increase generalization performance. 

\subsubsection{2D Appearance Features}
Each detection generates an appearance that is potentially observed on camera. For each detection, the 3D bounding box corners are projected into the image plane and a convex hull of that set is computed. A hull-enclosing rectangle defines the image patch and the respective camera showing most of the object is selected. Thus, the approach includes object backgrounds under the assumption that in-between frames the background stays approximately constant. A fully-convolutional auto-encoder architecture utilizing residual skip connections is employed to learn image features $\mathbf{h}_{\text{2D-A},j}$ as latent space representations. In the case of occlusions, we still use that respective appearance feature and overcome this issue using higher-order similarity finding through the chosen GNN architecture.

\subsubsection{3D Appearance Features}
In order to include 3D shape information, the sparse LiDAR point cloud within and in close proximity to the objects' 3D bounding box is extracted while neglecting the points' reflectance value. In order to account for pose estimation errors a slightly enlarged cuboid is used to associate LiDAR points to the respective object. The masked point cloud is encoded using a PointNet-architecture \cite{pointnet} that is trained towards predicting object categories. This is motivated by prior works that showed that PointNet works well on segmented point clouds \cite{gnn3dmot, pointnet}. A higher-dimensional feature $\mathbf{h}_{\text{3D-A},j}$ (128-dim.) is taken as the 3D-AL feature used for tracking.

\subsubsection{Radar Features}
In addition to LiDAR measurements, radar detections can be used for two reasons: Firstly, they provide a highly accurate radial velocity measurement between the particular sensor and the object (not the actual velocity) and secondly, they provide measurements of objects in large distances, which are captured imperfectly with cameras or sparse LiDAR readings. The measured radial velocity $v_r$ is split into two orthogonal components $(v_x, v_y)$ represented in the ego-vehicle frame that are each compensated by the ego-vehicle motion~\cite{nuscenes}. The raw set of radar reflections is clustered and the height coordinate is neglected because the radar's longitudinal wave characteristic renders the height coordinate not decisive and erroneous more often than not. We arrive at a radar parametrization $r_P = (x,y,v_x,v_y)$, where $x,y$ is the 2D object position after transformation from the radar coordinate frame into ego-vehicle coordinates. Since each radar detection does not hold a height coordinate we perform 2D pillar expansion~\cite{nabati2021centerfusion} and associate radar detections to objects as soon as the enlarged objects' cuboid and the pillar intersect. Since the chosen pillar representation does not represent an element of a Euclidean group as in the LiDAR case, we follow a naive approach and remove all coordinate-sensitive transforms in the PointNet architecture and merely transform each object's feature that consists of multiple radar detections in a permutation-invariant manner to arrive at the radar appearance feature $\mathbf{h}_{\text{R},j}$.\looseness=-1

\begin{figure*}
    \centering
    \includegraphics[width=1.0\textwidth]{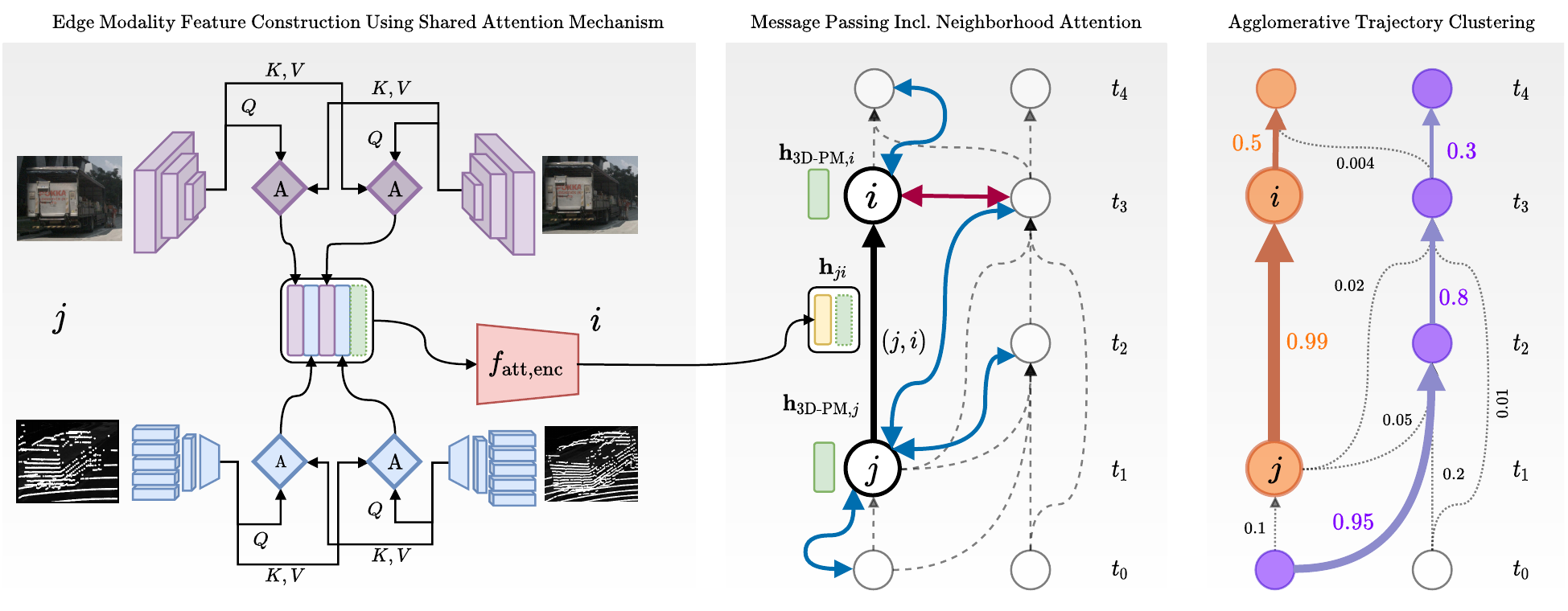}
    \caption{Overview of our Batch3DMOT architecture. A cross-edge modality attention mechanism fuses the features of the involved objects to construct an edge feature (left). Message passing including inter-category neighborhood attention propagates information. Blue arrows denote time-aware message passing, and red arrows denote frame-wise information propagation (middle). Predicted edge scores are turned into trajectory hypotheses using agglomerative trajectory clustering (right).}
    \label{fig:cross_edge_mod_att}
    \vspace{-.5cm}
\end{figure*}

\subsection{Graph Construction}

The chosen approach arranges nodes on a tracking graph over 5 frames in a sliding window manner with a stride of 1. The tracking performance improves drastically when learning the tracking task with actual detections instead of ground truth annotations since FP filtering and FN handling pose major challenges in real-world tracking. Consequently, ground truth annotation identifiers need to be paired with actual detections in order to construct edge labels for the learning stage. Under the assumption that ground truth annotations generally do not show significant intra-category overlap, we match detection results to geometrically close annotations in the birds-eye view (BEV).\looseness=-1

\subsubsection{Initial Node and Edge Embeddings}
The initial node features only consists of the 3D-PM feature:
\begin{equation}
    \mathbf{h}_{i} =  \left[\mathbf{h}_{\text{PM},i} \right]^T \in \mathbb{R}^{11+C+96}.
\end{equation}
Throughout this work, we found that it is more beneficial to add modality such as $\mathbf{h}_{\text{2D-A},i}$ during the message passing stage instead of as an initial feature as shown in \refsec{sec:experiments}. The edge features are defined as
\begin{equation}
    \mathbf{h}_{ji} = \left[\Delta x_{ji}, \Delta v_{ji}, \Delta \gamma_{ji}, \Delta s_{ji}, \Delta t_{ji} \right]^T \in \mathbb{R}^5 ,
\end{equation}
where $\Delta x_{ji}$ denotes Euclidean distance between the object centers and $\Delta v_{ji}$ the L2 norm of both velocity vectors. The smallest signed yaw difference of the two detections is given by $\Delta \gamma_{ji}$ while $\Delta s_{ji}$ is the log-volume-ratio and $\Delta t_{ji}$ the time difference.

\subsubsection{Graph Connectivity}
When investigating the effect of graph connectivity on the tracking result we found that it is beneficial to only connect nodes of the same object category rather than utilizing inter-category edges that could potentially overcome class prediction errors from the object detection task. This essentially renders the problem a disconnected graph with multiple components, which generally limits information propagation when learning. While the 2D MOT task is mostly focused on one object category, the 3D MOT task faces both the curse of dimensionality leading to a high number of detections per frame and multiple object categories. As a consequence, we found that limiting the number of possible edges represents an essential prior to the learning problem. Based on the normalized kinematic similarity metric
\begin{equation}
    v_{ji}^{*} = \frac{\frac{1}{2}\Delta x_{ji}^{*} + \frac{1}{4}\Delta \gamma_{ji}^{*}  + \frac{1}{4}\Delta v_{ji}^{*}}{\operatorname{max}_{q} \{\frac{1}{2}\Delta x_{qi}^{*} + \frac{1}{4}\Delta \gamma_{qi}^{*} + \frac{1}{4}\Delta v_{qi}^{*} \mid \forall k: t_{q}<t_i \}},
\label{eq:kinematic_sim_metric}
\end{equation}
the k-nearest neighbors in the past of every node $i$ are selected for edge construction, which essentially extracts a promising corridor based on similar position, velocity vectors, and yaw angles, while $\Delta x_{qi}^{*}, \Delta \gamma_{qi}^{*}, \Delta v_{qi}^{*}$ itself represent neighborhood-normalized distances. In the following, directed edges are constructed pointing from each of the $k$ neighbors to node $i$. Regarding the following graph learning step, edge labels denoting the active/inactive edges are necessary. We examine whether two nodes hold the same instance identifier and only connect them as an active edge if they represent the closest time-wise occurrence of the same object instance. Otherwise, edges hold labels of value 0.


\subsection{Message Passing Graph Neural Network}

This work employs the principle of time-aware neural message passing~\cite{braso2020learning}, which is extended to allow information exchange between inter-category nodes that reside in particular disconnected graph components. In addition, we present a novel way to include intermittent sensor modalities across edges. 

\subsubsection{Cross-Edge Modality Attention}
Initial node and edge features are encoded to produce approximately evenly-sized node and edge embeddings
\begin{equation}
    f_{enc}^{v}(\mathbf{X}) = \mathbf{H}_{v}^{(0)}, \quad f_{enc}^{e}(\mathbf{X}_{e}) = \mathbf{H}_{e}^{(0)},
\end{equation}
where the two networks take the form of MLPs. In the case of additional sensor modalities, the edge feature is augmented using modality cross-attention between the respective nodes' features to which edge (j,i) is incident to. Thus, each nodes' feature is attending and is being attended in order to find an agreement on whether to utilize the respective modality during the edge feature update. Based on that, we define the following queries $\mathbf{Q}$, keys $\mathbf{K}$ and values $\mathbf{V}$ for both attention directions:
\begin{equation}
    \mathbf{Q}_{ij}=\mathbf{X}_{\text{sens},i}\quad \mathbf{K}_{ij}=\mathbf{X}_{\text{sens},j}\quad \mathbf{V}_{ij}=\mathbf{X}_{\text{sens},j} 
\end{equation}
\begin{equation}
    \mathbf{Q}_{ji}=\mathbf{X}_{\text{sens},j} \quad
    \mathbf{K}_{ji}=\mathbf{X}_{\text{sens},i} \quad \mathbf{V}_{ji}=\mathbf{X}_{\text{sens},i},
\end{equation}
where $\mathbf{X}_{\text{sens},j}$ could represent either a LiDAR $\mathbf{h}_{\text{3D-A},j}$ or radar feature $\mathbf{h}_{\text{R},j}$ of the respective node. We use a standard multi-head attention mechanism per modality in order to compute attention-weighted features using head-specific linear transforms $(\mathbf{W}_{u}^{Q}, \mathbf{W}_{u}^{K}, \mathbf{W}_{u}^{V})$ to attend to multiple regions with the respective modality feature:
\begin{equation}
    \operatorname{MultiHead}(\mathbf{Q},\mathbf{K},\mathbf{V}) = \operatorname{Concat}(head_1,...head_h)\mathbf{W}^{O}, 
\end{equation}
\begin{equation}
    \text{where }\operatorname{head}_u = \operatorname{Softmax}(\frac{\mathbf{Q}\mathbf{W}_{u}^{Q})(\mathbf{K}\mathbf{W}_{u}^{K})^{T}}{\sqrt{d_k}})\mathbf{V}\mathbf{W}_{u}^{V}.
\end{equation}
The attended modality features are then concatenated and encoded as depicted in \reffig{fig:cross_edge_mod_att}. In the case of both LiDAR and camera we arrive at:
\begin{equation}
    \mathbf{H}_{e,\text{att}}^{*} = f_{\text{att,enc}}\left(\left[\mathbf{X}_{\text{3D-A},i}^{*}, \mathbf{X}_{\text{2D-A},i}^{*}, \mathbf{X}_{\text{3D-A},j}^{*}, \mathbf{X}_{\text{2D-A},j}^{*}, \mathbf{X}_{e} \right]\right),
\end{equation}
which constitutes the attention-weighted modality edge feature used during the edge update step in message passing that is briefly covered in the following.

\subsubsection{Time-Aware Message Passing Using Inter-Category Graph Attention}
A single message passing layer consists of an edge feature update based on the neighboring node features $\mathbf{h}_{i}^{(l-1)}$, $\mathbf{h}_{j}^{(l-1)}$ and the current edge feature $\mathbf{h}_{ji}^{(l-1)}$. In addition, our approach involves the multi-modal attention-weighted similarity feature $\mathbf{h}_{ji,\text{att}}^{(0)}$ , which is fed in each iteration to substantiate the update based on appearance similar modality features:\looseness=-1
\begin{equation}
\mathbf{h}^{(l)}_{ji} = f_{e}\left(\left[\mathbf{h}_{i}^{(l-1)},\mathbf{h}_{j}^{(l-1)},\mathbf{h}_{ji}^{(l-1)}, \mathbf{h}_{ji,\text{att}}\right]\right),
\end{equation}
where $[\cdot,\cdot,\cdot]$ represents the concatenation of the four representations as an input to a ReLU-activated MLP $f_{e}$. With respect to updating node features, messages $\mathbf{m}_{ij}^{(l)}$ are crafted based on either neighbors in the past $\mathcal{N}_{past}(j)$ or neighbors in the future $\mathcal{N}_{fut}(j)$ of a node $j$. In the next step, all messages from the future and from the past neighborhood of a node, respectively, are aggregated using a permutation-invariant sum, which results in node-specific past and future features:
\begin{equation}
\mathbf{h}_{j, past}^{(l)} = \sum_{i \in \mathcal{N}_{past}(j)} \underbrace{f_{v}^{past}\left(\left[\mathbf{h}_{i}^{(l-1)}, \mathbf{h}_{ji}^{(l)}, \mathbf{h}_{i}^{(0)}\right]\right)}_{\mathbf{m}_{ij}^{(l)}} ,
\end{equation}
\begin{equation}
\mathbf{h}_{j, fut}^{(l)} = \sum_{i \in \mathcal{N}_{fut}(j)} \underbrace{f_{v}^{fut}\left(\left[\mathbf{h}_{i}^{(l-1)}, \mathbf{h}_{ji}^{(l)}, \mathbf{h}_{i}^{(0)}\right]\right)}_{\mathbf{m}_{ji}^{(l)}}.
\end{equation}
The functions $f_{v}^{fut}$ and $f_{v}^{fut}$ again take the form of MLPs and transform the recently updated edge feature $\mathbf{h}_{ji}^{(l)}$, the initial node feature $\mathbf{h}_{i}^{(0)}$ as well as the current node representation $\mathbf{h}_{i}^{(l-1)}$. The final node update is reached by combining the past and future feature and feeding it to a function $f_v$:
\begin{equation}
\mathbf{h}_{j}^{(l)}=f_{v}\left(\left[\mathbf{h}_{j, past}^{(l)}, \mathbf{h}_{j, fut}^{(l)}\right]\right),
\end{equation}
which again takes the form of a ReLU-activated MLP. 

\begin{figure}
    \centering
     \includegraphics[width=1.0\linewidth]{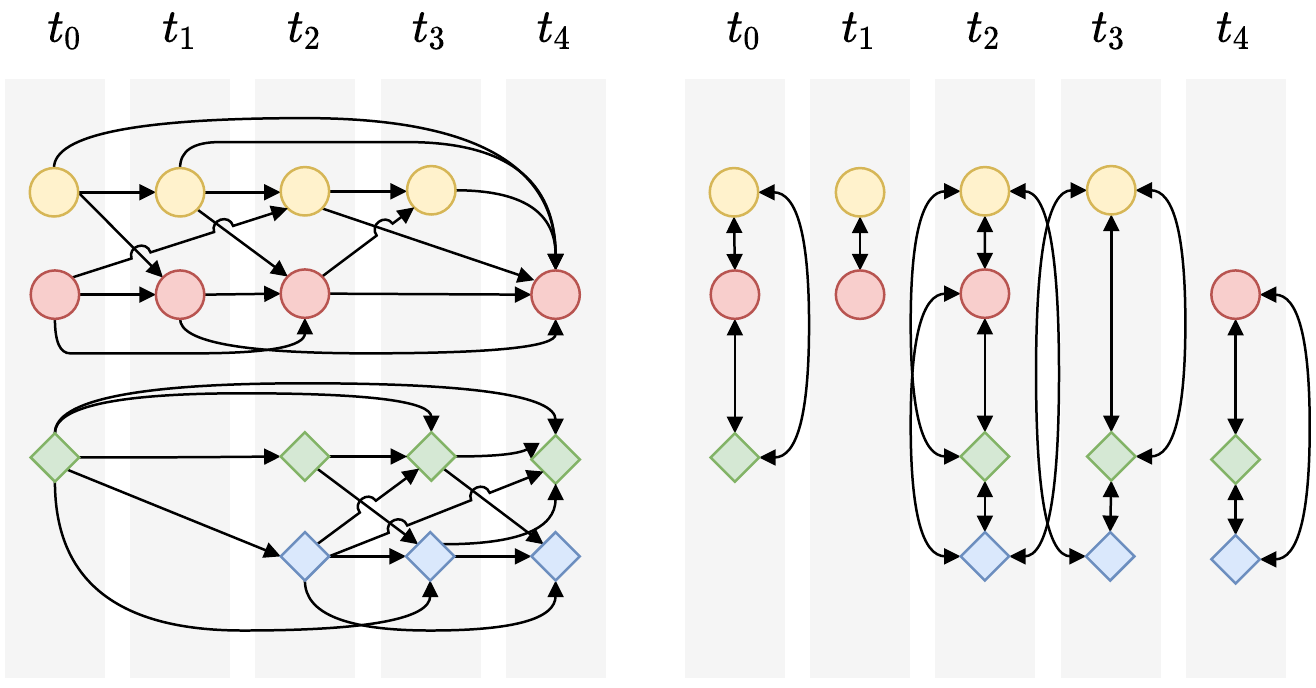}
     \caption{Both graph connectivity cases: Time-aware message passing operates on a time-directed acyclic tracking graph that holds disconnected semantic components (left) while attention-weighted neighborhood convolution is performed on temporary frame-wise k-NN graphs (right).}
     \label{fig:connectivity}
     \vspace{-0.5cm}
\end{figure}

In order to enable inter-category information exchange in between message passing steps, we construct temporary frame-wise graphs for which each node $j$ is connected to its top-k neighbors (without self-loops) having identical timestamps regardless of object category. Then, a graph attention layer (GAT)~\cite{velivckovic2018graph} propagates node features in an attention-weighted manner to produce linear combinations of neighbor nodes:
\begin{equation}
    \mathbf{h}'_{j} = \alpha_{ii}\mathbf{\Theta} \mathbf{h}_{j}^{(l)} + \sum_{i \in \mathcal{N}_{t}(j)}{ \alpha_{ji}\mathbf{\Theta} \mathbf{h}_{i}^{(l)}},
\end{equation}
while $i$ and $j$ in this case do not represent nodes in different frames but in an identical one ($t_i = t_j$). Attention weights among the nodes per frame are used to elect nodes that are of relevance to one another such as overlapping detections of different semantic categories, which would normally reside in two different disconnected graph components. The attention weights\looseness=-1
\begin{equation}
    \alpha_{ji}=\frac{\exp \left(\operatorname{LeakyReLU}\left(\mathbf{a}^{T}\left[\mathbf{W} \mathbf{h}_{i} ,\mathbf{W} \mathbf{h}_{j}\right]\right)\right)}{\sum_{k \in \mathcal{N}_{i}} \exp \left(\operatorname{LeakyReLU}\left(\mathbf{a}^{T}\left[\mathbf{W} \mathbf{h}_{i} ,\mathbf{W}\mathbf{h}_{k}\right]\right)\right)}
\end{equation}
are softmax-normalized across neighborhoods while the attention-mechanism consists of a single-layer feedforward neural network represented by a weight vector $\mathbf{a}$.
A final edge classifier MLP $f_e^{reg}$ downprojects each edge feature to a single Sigmoid-activated scalar that denotes the edge activation score.

\subsubsection{Loss Formulation}
As the approach considers multiple semantic categories that exhibit different frequencies of occurrence, it faces significant category imbalance with respect to the number of nodes per category. This directly translates to an even more unbalanced category-specific number of edges contained in the graph. Having chosen a disconnected graph, the nodes incident to a particular edge are always of the same category. Based on that, we employ a class-balanced loss formulation that takes a binary cross-entropy and weights edges based on category frequencies:
\begin{equation*}
    \mathcal{L}_{\text{CB}} = \frac{1}{\vert E \vert} \sum_{(j,i)\in E} \frac{1-\beta}{1-\beta^{n_{ji}}} y_{ji}\operatorname{log}(p_{\phi}^{ji})+(1-y_{ji})\operatorname{log}(1-p_{\phi}^{ji})\label{eq:cb},
\end{equation*}
where $\beta$ represents a hyperparameter and $n_{ji}$ is the absolute number of objects with respect to the node categories involved per edge. The respective weights are estimated based on object category frequencies in the training set~\cite{cui2019class}.

\subsection{Inference and Graph Traversal}
The outputs of the GNN architecture are Sigmoid-valued scores that represent whether an edge is likely to be active/inactive. Instead of thresholding at an edge score of 0.5 to find active/inactive edges to turn into trajectories, we follow the spirit of \textit{ByteTrack}~\cite{bytetrack} and try to associate (nearly) \textit{every} detection with a preliminary trajectory. Based on the assumption that the predicted edge scores show some inherent order, i.e., FP edges exhibit lower scores than TP edges within local neighborhoods of the graph, we propose a score-based agglomerative trajectory clustering paradigm (\refalg{alg:trajclustering}). The edge score predictions of multiple overlapping batches are averaged per edge. All edges are arranged in descending order and empty (ordered) clusters are initialized that will later hold output trajectories. In the following, we loop through all edges from the highest to lowest score and check whether the edge is constrained or unconstrained. If constrained, it is checked whether the edge would essentially add time-wise leading or trailing nodes to one of the temporary clusters or if it joins two clusters. In the case of joining two clusters, an additional score-wise threshold needs to be met. Otherwise, the edge does not violate any tracking constraints and a new cluster is initialized.\looseness=-1
\begin{algorithm}[t]
    \footnotesize
    \SetAlgoLined
    \SetAlgoNoEnd
    $E_{pred}, N_{meta} \leftarrow $ CombineBatches($GNN(\mathbf{X},\mathbf{X}_e)$)\\
    $E_{pred}^{*} \leftarrow $ DescSortEdgesByScore(${e_{pred}}$)\\
    $vis \leftarrow$ CreateVisitedNodesDict()\\
    $\mathcal{C} \leftarrow$ CreateEmptyClustersDict() \\
    \For{$e_{ji},score$ in $E_{pred}^{*}$ }{
        \eIf{$j \notin vis$ and $i \notin vis$}{
            $\mathcal{C}\leftarrow$ CreateNewCluster($e_{ji}$) \\
            UpdateVisitedNodes($e_{ji}, \mathcal{C}$) \\
        }{
            \uIf{$j \notin vis$ and $i \in vis$}{
                \If{ $i$ is leading node in C($i$)}{
                    $\mathcal{C}\leftarrow$ AddToCluster($e_{ji}$) \\
                    $vis \leftarrow$ UpdateVisitedNodes($e_{ji}, \mathcal{C}$) \\
                }
            }
            \uElseIf{$j \in vis$ and $i \notin vis$}{
                \If{ $j$ is trailing node in $\mathcal{C}(j)$}{
                    $\mathcal{C}\leftarrow$ AddToCluster($e_{ji}$) \\
                    $vis \leftarrow$ UpdateVisitedNodes($e_{ji}, \mathcal{C}$) \\
                }
            }
            \uElseIf{$j \in vis$ and $i \in vis$}{
                \If{$j$ is trailing $\mathcal{C}(j)$ and $i$ is leading $\mathcal{C}(i)$}{
                    $\mathcal{C}\leftarrow$ JoinClusters($e_{ji}$) \\
                    $vis \leftarrow$ UpdateVisitedNodes($e_{ji}, \mathcal{C}$) \\
                }
            }
        }
    }
    \textbf{return} TurnClustersIntoTrajectories($\mathcal{C}$)
    \caption{Agglomerative Trajectory Clustering.}
    \label{alg:trajclustering}
\end{algorithm}

\section{Experimental Evaluation}
\label{sec:experiments}

In this section, we present quantitative and qualitative evaluations of our proposed Batch3DMOT on the nuScenes~\cite{nuscenes} and KITTI~\cite{kitti} datasets using the average multiple-object tracking accuracy (AMOTA) and multiple-object tracking accuracy (MOTA) metrics, respectively. 
Similar to existing methods, we evaluate our model on the nuScenes test set as well as the KITTI 2D MOT benchmark. We provide additional experimental data in the supplementary material.


{\parskip=5pt
\noindent\textit{Detections and GT Matching}: In this approach, we use the detections provided by MEGVII~\cite{megvii_cbgs} and CenterPoint~\cite{centerpoint} for nuScenes. On the KITTI dataset, we use Point-RCNN detections \cite{shi2019pointrcnn} as also used by FG3DMOT~\cite{poschmann2020factor} and AB3DMOT~\cite{ab3dmot}. We match detections to ground truth trajectory labels to obtain identifiers. As proposed earlier~\cite{megvii_cbgs, nuscenes}, the L2 center distance is often used for matching, which is beneficial for faraway objects. Our empirical findings show that especially large objects suffer from this heuristic since, e.g., their respective length is not predicted correctly, which leads to considerable object center translations and effectively renders the L2 distance uninformative. Therefore, we follow a bi-level approach by first selecting a close radius (L2) and then checking whether detection and annotation exhibit a significant BEV-IoU overlap.}

\begin{table}
    \footnotesize
    \centering
    \caption{Comparison of AMOTA scores on the nuScenes validation set. Bold/underlined numbers denote \textbf{best}/\underline{second best} model scores, respectively.}
    \label{tab:categ_val}
    \tabcolsep=0.11cm
    \resizebox{\linewidth}{!}{
        \begin{tabu}{ l | c | c c c c c c c}
        \toprule
        Method & Overall & Bicyc. & Bus & Car & Moto. & Ped. & Trailer & Truck \\
         \midrule
         \textit{AB3DMOT \cite{ab3dmot}\cite{megvii_cbgs}}  & 0.179 & 0.09 & 0.489 & 0.36 & 0.051 &  0.091 &  0.111 & 0.142 \\
         \textit{Prob3DMOT \cite{chiu2020probabilistic}\cite{megvii_cbgs}} & 0.561 & 0.272 & 0.741 & 0.735 & 0.506 & 0.755 & 0.337 & 0.580 \\
         CenterPoint \cite{centerpoint} & 0.665 & 0.458 & 0.801 & 0.842 & 0.615 & 0.777 & \underline{0.504} & 0.656 \\
         ProbMM-3DMOT \cite{chiu2021probabilistic} & 0.687 & 0.490 & 0.820 & 0.843 & 0.702 & 0.766 & \textbf{0.534} & 0.654 \\
         \midrule
         3D-PM-MEGVII\cite{megvii_cbgs}  & 0.623 &  0.368 & 0.759 & 0.789 & 0.655 & 0.796 & 0.378 & 0.617 \\
         3D-PM-CP \cite{centerpoint} & 0.708 & 0.540 & 0.837 & 0.849 & 0.728 & 0.813 & 0.497 & 0.689 \\
         3D-PM-C-CP \cite{centerpoint}  & 0.709 & \underline{0.542} & 0.837 & \textbf{0.851} & 0.733 & 0.813 & 0.502 & 0.688   \\
         3D-PM-CL-CP \cite{centerpoint}  & \textbf{0.715} & 0.540 & \textbf{0.855} & \textbf{0.851} &  \textbf{0.748} & \textbf{\underline{0.821}} & 0.493 & \underline{0.695} \\ 
         3D-PM-CLR-CP \cite{centerpoint} & \underline{0.713} & \textbf{0.545} & \underline{0.851} & \underline{0.850} & \underline{0.736} & \underline{0.820} & 0.494 & \textbf{0.696} \\ 
        \bottomrule
        \end{tabu}
    }
    \vspace{-.5cm}
\end{table}

\begin{table*}
    \scriptsize
    \centering
    \caption{Ablation study on the nuScenes validation set. All results shown are derived using CenterPoint detections \cite{centerpoint}.}
    \vspace{1mm}
    \begin{tabu}{l |c | c | c | c | c c c c c c c c }
    \toprule
        Method  & PM & C & L & R & AMOTA$\uparrow$ & AMOTP$\downarrow$ & MOTA$\uparrow$ & Recall$\uparrow$ & FP$\downarrow$ & FN $\downarrow$ & IDS $\downarrow$ & FRAG $\downarrow$ \\
        \midrule 
        CenterPoint \cite{centerpoint}  & \checkmark & & & & 0.664 & \textbf{0.567} & 0.562 & 0.698 & 13187 & 20446 & \underline{562} & 424  \\

        
        
        
        
        OGR3MOT \cite{zaech2021learnable}  & \checkmark & & & & 0.693 & 0.627 & 0.602 & -- & -- & -- & \textbf{262} & \textbf{332}  \\

        \midrule
        w/o MP layers  & \checkmark &  &  & & 0.519 & 0.960 & 0.471 & 0.592 & \textbf{7206} & 33801  & 7065 & 2648 \\
        60 k-NN  & \checkmark &  &  & & 0.578 & 0.728 & 0.493 & 0.633 & 12159 & 27187 & 4497 & 1646 \\ 
        Connected graph comp.  & \checkmark & \checkmark & \checkmark & & 0.646 & 0.842 & 0.599 & 0.702 & \underline{7621} & 23011  & 1233 & 761 \\ 
        TA-NMP \cite{braso2020learning}  & \checkmark & \checkmark & & & 0.668 & 0.698 & 0.589 & 0.714 & 11106 & 21806  & 1810 & 769 \\ 
        w/o Aggl. Traj. Clust. &  \checkmark & & & & 0.683 & 0.682 & 0.592 & 0.699 & 11030 & 20260 & 1271 & 434 \\
        Stacked modalities & \checkmark & \checkmark & \checkmark & & 0.689 & 0.678 & 0.602 & 0.688 & 9525 & 20954  & 938 & 536 \\
        w/o 2D-A attention & \checkmark & \checkmark & & & 0.698 & 0.657 & 0.602 & 0.700 & 9951 & 20641 & 886 & 403 \\
        w/o CB Loss & \checkmark &  &  & & 0.702 & 0.617 & 0.607 & \underline{0.723} & 11467 & \underline{18516} & 758 & 418 \\
        w/o Neigborhood GAT & \checkmark & & & & 0.703 & 0.644 & 0.604 & 0.716 & 11465 & 18691  & 767 & 416 \\
        \midrule
        Batch3DMOT-3D-PM &  \checkmark & & & & 0.708 & 0.630 & \textbf{0.612} & 0.719 & 11102 & 18640  & 688 & 383 \\
        Batch3DMOT-3D-PM-C &  \checkmark &\checkmark & & & 0.709 & 0.622 & 0.608 & 0.716 & 11307 & 18722 & 664 & 375 \\
        \textbf{Batch3DMOT-3D-PM-CL} &  \checkmark & \checkmark & \checkmark& & \textbf{0.715} & 0.598 & \textbf{0.612} & \textbf{0.726} & 11175 & \textbf{18494} & 598 & \underline{357} \\
        Batch3DMOT-3D-PM-CLR &  \checkmark & \checkmark & \checkmark& \checkmark & \underline{0.713} & \underline{0.592} & \underline{0.611} & \textbf{0.726} & 11196 & 18520 & 622 & 385 \\ 
         \bottomrule
    \end{tabu}
    \vspace{-2mm}
    \label{tab:ablation_val}
\end{table*}

\begin{figure*}
\centering
\footnotesize
\setlength{\tabcolsep}{0.1cm}
\begin{tabular}{P{5.8cm}P{5.8cm}P{5.8cm}}
\includegraphics[width=\linewidth,height=4.2cm]{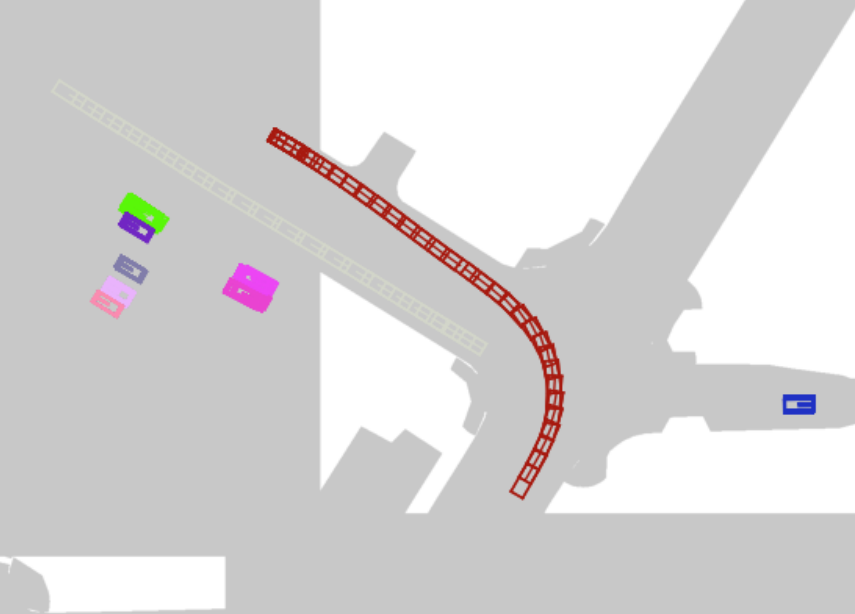} & \includegraphics[width=\linewidth,height=4.2cm]{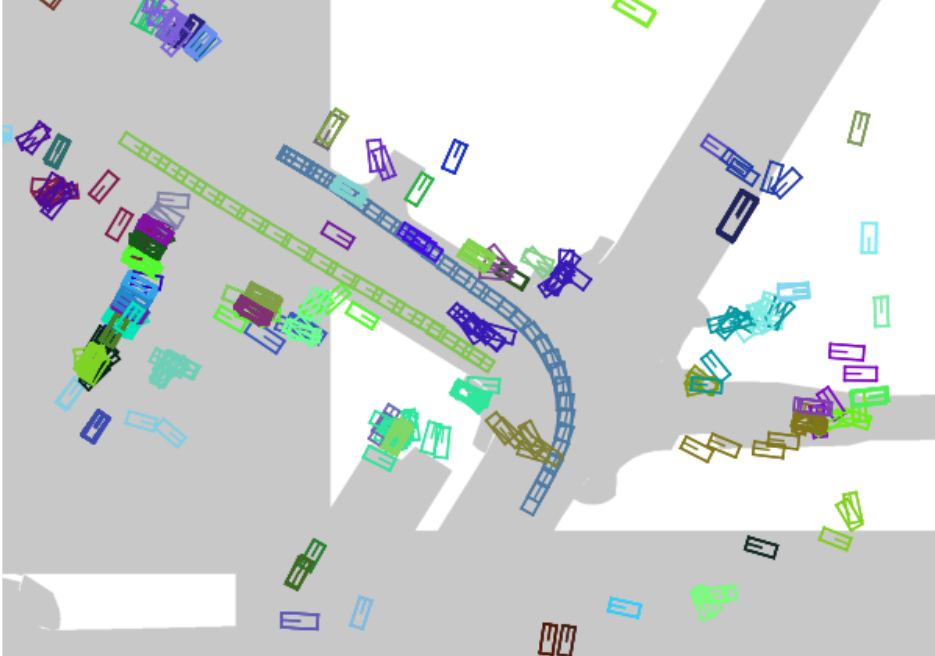} & \includegraphics[width=\linewidth,height=4.2cm]{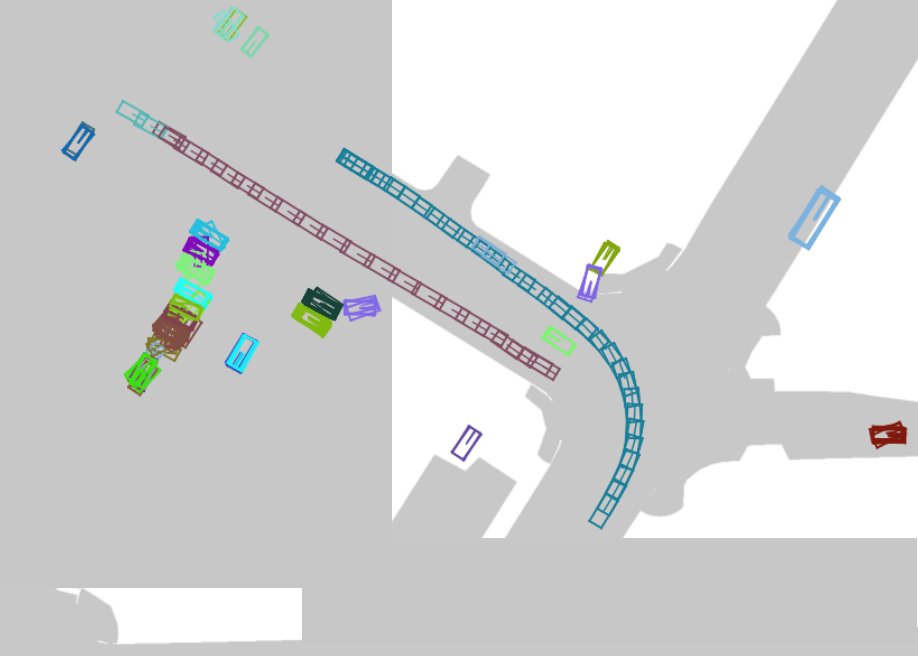} \\
(a) Ground Truth Trajectories & (b) Kalman Filter-Based Tracking \cite{chiu2021probabilistic} & (c) Batch3DMOT (Ours) \\
\end{tabular}
\caption{Comparison of FP filtering on the nuScenes validation set. The Kalman filter-based tracking approach \cite{chiu2021probabilistic} shows mediocre FP filtering, while our proposed Batch3DMOT using only 3D-PM features for pseudo ground truth generation shows superior FP filtering.} 
\label{fig:qualitative}
\vspace*{-0.5cm}
\end{figure*}

{\parskip=5pt
\noindent\textit{Implementation Details}: Each batch consists of five frames where each node is connected to its 40-nearest neighbors in the past. Object 2D-A features are scaled to a $32\times32$-dimensional RGB image. The fully-convolutional image encoder is built upon the ResNet architecture and is trained for $80$ epochs using a learning rate (LR) of $0.002$ and a batch size of 32. LiDAR point clouds are aggregated over multiple frames, normalized, centered, and rotationally permuted to mimic orientation errors in 3D object detection. In order to generate a $\text{3D-A}$ feature, a minimum of five LiDAR points must exist. Otherwise, the object does not hold a 3D-A feature. The LiDAR PointNet is trained for 500 epochs using a batch size of 64 and a LR of $0.001$. We do not employ pre-training on other datasets contrary to previous findings~\cite{braso2020learning} since this decreased model performance. In the case of radar, each object must hold at least two radar detections to generate a feature. The radar network is trained for 1000 epochs using a batch size of 256 and a LR of 0.0002. The GNN models are trained for 100 epochs using a batch size of two (10 frames in total) and LRs between $4e^{-5}$ and $1e^{-4}$, which largely depends on the detections used and the number of edges contained in the graph. \rebuttal{We perform 6 message passing steps with intermediate frame-wise neighborhood convolutions (20 k-NN), which we ablate in the experiments presented in the supplementary material in \secref{sec:additional_ablation}.C.} We estimate the class-balancing factors based on the absolute frequencies of ground truth annotations in the training set and find the hyperparameter $\beta=0.8$ empirically. \rebuttal{While the modality attention modules use two attention heads, it proved to be sufficient to use a single head in the frame-wise neighborhood attention mechanism. Our evaluations show that training feature encoders and the GNN model in an end-to-end manner or using transfer learning decreases the performance.}
}

\subsection{Quantitative Results and Ablation Study}
We report our results on the nuScenes validation set in \tabref{tab:ablation_val}. Additionally, we also present the category-specific AMOTA scores in \tabref{tab:categ_val}. While there is no existing offline method on the nuScenes benchmark, we compare against a strong set of state-of-the-art online trackers. The main baselines include AB3DMOT~\cite{ab3dmot}, Prob3DMOT~\cite{chiu2020probabilistic}, two online Kalman filter-based methods, and CenterPoint~\cite{centerpoint} which performs closest-point matching of predicted velocity vectors. These methods currently represent the naive choice when generating pseudo ground truth due to their inherent simplicity and robustness as well as high tracking accuracy in terms of the AMOTA score. Nonetheless, we argue that there is room for improvement by examining multiple frames in a batch-manner. In our case, we chose a batch length of 5 frames for three reasons: 
\begin{enumerate*}
    \item Typical birth and death memory matching time thresholds used in bipartite association~\cite{ab3dmot} are in a similar range.
    \item We expect an object to reappear after 2-3 frames of false negative detections while neglecting long-term occlusions.
    \item With an increasing number of objects per batch, the number of edges increases exponentially, which makes the graph learning problem significantly more complex.
\end{enumerate*}
Based on these factors, we see the grounds for comparison with the chosen baselines. \rebuttal{We ablate on the number of frames to consider in \secref{sec:additional_ablation}.A. of the supplementary material.}

\tabref{tab:ablation_val} shows that an increase from 40 to 60 k-nearest neighbors per node results in a stark decrease in tracking accuracy (-12.8\%). The best choice of $k$ can only be found empirically, however $10e^{3}$ serves as a suitable maximum number of edges per batch. \rebuttal{\figref{fig:param_study}~(b) in the supplementary material presents the tracking performance for 10 and 20 nearest neighbors.} Independently, we observe a considerable decrease in tracking performance when connecting nodes of different semantic categories (-6\%). Phrasing the offline 2D MOT GNN by Braso~\textit{et~al.}~\cite{braso2020learning} as an offline 3D MOT method provides an additional baseline. It uses identical edge features but the 2D-A feature as the sole node feature, which performs worse than our architecture, but achieves similar recalls. Moreover, reducing our model to a node similarity network (no message passing) results in an AMOTA score of 0.519. Furthermore, we observe a slight decrease in AMOTA when the frame-wise neighborhood GAT aggregations are removed (-0.5\% wrt. best performing model). Modality intermittence is especially severe when stacking all the modalities as a node feature (\tabref{tab:ablation_val}), which ultimately motivated our modality attention mechanism. Furthermore, we also observe that the class-balancing scheme slightly enhances the tracking result. Finally, we test-wise replace our agglomerative trajectory clustering with a bidirectional depth-first-search algorithm that iterates from high score to low score edges, which performs worse than our proposed agglomerative clustering paradigm. \rebuttal{We provide a more extensive parameter study in \secref{sec:additional_ablation} of the supplementary material.}

\rebuttal{We observe the highest AMOTA tracking score for the 3D-PM-CL model on the validation set (see \tabref{tab:ablation_val}), which demonstrates the efficacy of the proposed modality attention module (see \tabref{tab:ablation_val}). Including 2D-A features leads to a small performance increase compared to the pose-only variant (3D-PM). Here, occlusions are the limiting factor that restricts a larger improvement in performance, which is supported by a large accuracy increase when including LiDAR. The PointNet architecture succeeds effectively at \textit{extracting} local object information. The 3D-PM-CLR architecture exhibits a slight accuracy decrease compared to the CL-counterpart, which can be attributed to the severe sparsity and quality of the radar detections in nuScenes. Additionally, we present the category-specific AMOTA scores in \tabref{tab:categ_val}. On the nuScenes test set, the 3D-PM-CL model outperforms the pose-and-motion variant (\tabref{tab:test}). Batch3DMOT achieves an AMOTA score of 0.689, which outperforms the state-of-the-art online method OGR3MOT~\cite{zaech2021learnable} by 3.3\% using the same detections. Note that EagerMOT uses both 2D and 3D detections which the other methods do not. We also report the performance on the KITTI test set for the car category in \tabref{tab:kitti_test} and observe that our model achieves competitive results as FG-3DMOT~\cite{poschmann2020factor}, while using only 5 frames.}
\begin{table*}
    \scriptsize
    \centering
    \caption{Comparison of the 3D MOT performance on the nuScenes test set evaluated on the official benchmarking server.}
    \vspace{1mm}
    \begin{tabu}{l l |c | c | c | c | c c c c c c c c }
    \toprule
         Method & Detections & PM & C & L & AMOTA$\uparrow$ & AMOTP$\downarrow$ & MOTA$\uparrow$ & Recall$\uparrow$ & FP$\downarrow$ & FN$\downarrow$ & IDS$\downarrow$ & FRAG$\downarrow$ \\
         \midrule
         \textit{AB3DMOT \cite{ab3dmot}} & \textit{MEGVII \cite{megvii_cbgs}} & \checkmark & &  & 0.151 & 1.501 & 0.154 &  0.276 & 15088 &  75730 &  9027 & 2557 \\
         \textit{Prob3DMOT \cite{chiu2020probabilistic}}  & \textit{MEGVII \cite{megvii_cbgs}} & \checkmark & & & 0.550 & 0.798 & 0.459 & 0.600 & 17533 & 33216 & 950 & 776 \\
         ProbMM3DMOT\cite{chiu2021probabilistic}  & CP \cite{centerpoint} & \checkmark & \checkmark & & 0.655 & 0.617 & 0.555 & \textbf{0.707} & 18061 & 23323 & 1043 & 717 \\
         CenterPoint \cite{centerpoint}  & CP \cite{centerpoint} & \checkmark & & & 0.650 & \textbf{0.535} & 0.536 & 0.680 & 17355 & 24557 & \underline{684} & 553  \\
         OGR3MOT \cite{zaech2021learnable}  & CP \cite{centerpoint} & \checkmark & & &0.656 & 0.620 & 0.554 & 0.692 & 17877 & 24013 & \textbf{288} & \textbf{371}  \\
         \textit{EagerMOT}~\cite{eagermot}  & \textit{CP~\cite{centerpoint} + Cascade RCNN}  & \checkmark & \checkmark & & 0.677 & 0.550 & 0.568 & 0.727 & 17705 & 24925 & 1156 & 601 \\
         
         \midrule
         Batch3DMOT (Ours) & CP \cite{centerpoint} & \checkmark & & & \underline{0.683} & 0.633 & \underline{0.568} & 0.679 & \textbf{15290} & \underline{22692} & 994 & 562 \\
         Batch3DMOT (Ours) & CP \cite{centerpoint} & \checkmark & \checkmark & \checkmark & \textbf{0.689} & \underline{0.604} & \textbf{0.570} & \underline{0.704} & \underline{15580} & \textbf{22353} & 718 & \underline{427} \\
         \bottomrule
    \end{tabu}
    \vspace{1pt}\\
     \rebuttal{We do not highlight \textit{methods that use different sets of detections \cite{ab3dmot, chiu2020probabilistic, eagermot}} but still report them in this table for completeness.}
    \vspace*{-.5cm}
    \label{tab:test}
\end{table*}

\subsection{Qualitative Results}

\figref{fig:qualitative} illustrates the accumulation over 40 frames of a scene on the nuScenes dataset. We observe that our proposed Batch3DMOT method removes a large number of detections that essentially represent FPs when compared with the trajectory ground truth. We also observe that our approach works especially well on still-standing objects, while Prob3DMOT~\cite{chiu2020probabilistic} yields a higher number of FPs. \rebuttal{Additional insight into low- and high-confidence model predictions is presented in \secref{sec:pseudo_traj_labels}, \secref{sec:qualitative_insights_suppl}, and \figref{fig:suppl_qual} of the supplementary material. \figref{fig:qualitative} meets the requirements with respect to generating pseudo-groundtruth. This is further exemplified in \secref{sec:pseudo_traj_labels}.B of the supplementary material by training on pseudo-labeled test-data and in \secref{sec:pseudo_traj_labels}.C for training an online 3D Kalman filter from data statistics that include weak pseudo-labeled annotations.}
\begin{table}
    \footnotesize
    \centering
    \caption{Comparison of the 2D MOT performance on the KITTI test set.}
    \label{tab:kitti_test}
    \tabcolsep=0.11cm
    \resizebox{\linewidth}{!}{
        \begin{tabu}{ l | c  c c c c c} 
        \toprule
        Method & MOTA$\uparrow$ & MOTP$\uparrow$ & MT$\uparrow$ & ML$\downarrow$ & IDS$\downarrow$ & FRAG$\downarrow$ \\
        \midrule
         \textit{DSM \cite{frossard}} & 0.762 & 0.834 & 0.600 & 0.831 & 296 & 868 \\
         AB3DMOT \cite{ab3dmot} & 0.838  & \underline{0.853} & 0.669 & 0.114 & \textbf{9} &  224 \\
         FG-3DMOT (online) \cite{poschmann2020factor} &  0.837 & 0.846 & 0.680 &  \underline{0.099}  &  \textbf{9} & 375  \\
         FG-3DMOT (offline) \cite{poschmann2020factor} & \underline{0.880} & 0.850 & \underline{0.755} &  0.119  &  20 & \underline{117} \\
         \midrule
         Batch3DMOT (5 frames) \cite{shi2019pointrcnn}   & \textbf{0.886} &  \textbf{0.868} & \textbf{0.767} & \textbf{0.088} & \underline{19} & \textbf{74} \\
        \bottomrule
        \end{tabu}
    }
    \vspace{-.2cm}
\end{table}

\section{Conclusion}

\rebuttal{In this work, we proposed a framework for addressing the offline 3D MOT task using a multi-modal graph neural network including a novel agglomerative trajectory construction scheme. We presented extensive results on two challenging datasets demonstrating that our approach achieves state-of-the-art performance. We also showed the benefits of our proposed cross-edge modality attention in mitigating the effect of modality intermittence. Our method was able to improve tracking accuracy compared to current online methods using the same detections and shows enhanced false positive filtering. In future work, we plan to extend our approach to also cope with long-term occlusions.}\looseness=-1

\typeout{}
\footnotesize
\bibliographystyle{IEEEtran}
\bibliography{references.bib}

\clearpage
\renewcommand{\baselinestretch}{1}
\setlength{\belowcaptionskip}{0pt}

\begin{strip}
\begin{center}
\vspace{-5ex}
\textbf{\LARGE \bf
3D Multi-Object Tracking Using Graph Neural Networks with\\\vspace{0.5ex}Cross-Edge Modality Attention} \\
\vspace{3ex}

\Large{\bf- Supplementary Material -}\\
\vspace{0.4cm}
\normalsize{Martin B\"uchner and Abhinav Valada}
\end{center}
\end{strip}

\setcounter{section}{0}
\setcounter{equation}{0}
\setcounter{figure}{0}
\setcounter{table}{0}
\setcounter{page}{1}
\makeatletter

\renewcommand{\thesection}{S.\arabic{section}}
\renewcommand{\thesubsection}{S.\arabic{subsection}}
\renewcommand{\thetable}{S.\arabic{table}}
\renewcommand{\thefigure}{S.\arabic{figure}}


\let\thefootnote\relax\footnote{Department of Computer Science, University of Freiburg, Germany.\\
Project page: \url{http://batch3dmot.cs.uni-freiburg.de}
}%

\normalsize

In this supplementary material, we (i) portray applications that delineate the usability of generated pseudo-labels, (ii) we present additional ablation experiments on main model parameters and (iii) give qualitative insights into low and high confidence model predictions.

\section{Generating Pseudo Trajectory Labels}
\label{sec:pseudo_traj_labels}

In this section, we describe different methodologies to demonstrate whether the proposed model can approximately meet demands deriving from reference-data generation.

\subsection{Trajectory Postprocessing}
\label{sec:postprocessing}

Since the proposed model does not include an update step fusing predictions and measurements, the predicted trajectories are estimates based on original detections. In order to yield smoother trajectories serving as labels, we further process them in a series of steps described below:
\begin{itemize}
  \item Trajectory interpolation regarding missing timesteps.
  \item Yaw angle projection into $\gamma \in [-\pi,\pi]$ range to prevent further interpolation errors.
  \item Compute an intra-track BEV-IoU as a measure for still-standing objects (e.g. parking cars), which is computed as the product of all BEV-IoUs of pairs of boxes.
  \item Yaw correction on still-standing objects that suffer from orientation error of approx. $\pm\pi$: For intra-track BEV-IoUs greater than 0.7, we cluster yaw angles into two regimes (track-wise). All angles contained in the track are overwritten using the mean yaw angle of the majority class.
  \item Due to the chosen batch-size of 5 trajectories the trajectories suffer from ID switches occasionally, which is convenient to detect for still-standing objects. For pairs of trajectories where each track shows an intra-track BEV-IoU $> 0.7$ we check for a BEV-IoU $> 0.6$ of the involved mean object poses of the two still-standing instances. If that threshold is met, the two trajectories are joined under one ID.
  \item Lastly, we interpolate trajectories using a weighted running average scheme in order to yield smoother object motion, which should ultimately guarantee a more suitable pseudo-ground truth.
\end{itemize}

Based on these trajectories, we conducted two additional experiments outlined in the following.

\begin{table}
    \footnotesize
    \centering
    \caption{Comparison of different training sets used to estimate Kalman filter covariance matrices of Prob3DMOT~\cite{chiu2021probabilistic} using CenterPoint~\cite{centerpoint} object proposals. Results are shown in terms of the AMOTA scores on the nuScenes validation set.}
    \label{tab:pseudo_prob3dmot}
    \tabcolsep=0.11cm
    \resizebox{\linewidth}{!}{
        \begin{tabular}{ r | c  c c c c c c c} 
        \toprule
        Training Set &  Overall & Bicyc. & Bus & Car & Moto & Ped. & Trailer & Truck \\
        \midrule
         \textit{nusc-train} & 0.614 & 0.387 & 0.791 & \textbf{0.780} & 0.528 & \textbf{0.698} & \textbf{0.494} & 0.622 \\
         \textit{nusc-train + pseudo-test} & \textbf{0.624}  & \textbf{0.436} & 0.808 & 0.779 & \textbf{0.549} & 0.693 & 0.457 & \textbf{0.645} \\
         \textit{pseudo-train + pseudo-test} & 0.611 & 0.377 & \textbf{0.822} & 0.768 & 0.540 & 0.695 & 0.447 & 0.625 \\
        \bottomrule
        \end{tabular}
    }
\end{table}

\begin{table*}
    \scriptsize
    \centering
    \caption{Comparison of different pseudo-label training schemes utilizing the 3D-PM model architecture on the nuScenes validation set.}
    \vspace{1mm}
    \begin{tabu}{r | c  | c  c  c  c  c c c }
    \toprule
        Training Set &  AMOTA$\uparrow$ & AMOTP$\downarrow$ & MOTA$\uparrow$ & Recall$\uparrow$ & FP$\downarrow$ & FN$\downarrow$ & IDS$\downarrow$ & FRAG$\downarrow$ \\
        \midrule
         \textit{nusc-train} & 0.708 & 0.630 & 0.612 & 0.719 & 11102 & 18640  & 688 & 383 \\
         \textit{nusc-train + pseudo-test} & 0.709  & 0.605 & 0.611 & 0.717 & 11470 & 18566 & 626 & 369 \\ 
         \textit{nusc-train + pseudo-test-entropy} & \textbf{0.711}  & 0.611 & 0.607 & \textbf{0.726} & 11323 & 18460 & 663 & 386 \\ 
         \textit{pseudo-train + pseudo-test} & 0.708 & 0.592 & 0.606 & 0.712 & 11958 & 18543 & 683 & 365 \\ 
         \textit{pseudo-test} & 0.705 & 0.592 & 0.605 & 0.722 & 12330 & 18317 & 724 & 376 \\ 
        \bottomrule
    \end{tabu}
    \label{tab:pseudo_training}
\end{table*}

\subsection{Pseudo-Label Training}
\label{sec:pseudo_label_training}

For testing the efficacy of our model predictions, we employ a pseudo-label training scheme that is exemplified for the nuScenes dataset. We use the additional data samples in the test split, which does not contain openly accessible annotations. We employ the 3D-PM-CL instance (see \tableref{tab:ablation_val}) optimized on the training set to yield pseudo-labels for the test-split. The postprocessing steps outlined in the previous section are applied to yield refined trajectories. Using a combination of the training split and pseudo-labels of the test-split (or a subset), a new 3D-PM model is trained. In general, we assume that the unlabeled test split is created by the same data generation process as the labeled share of the training dataset.

\begin{figure*}
\centering
\footnotesize
\begin{tabular}{P{9cm}P{9cm}}
\includegraphics[width=\linewidth, trim={0.5cm 0.5cm 0.5cm 0.5cm}, clip] {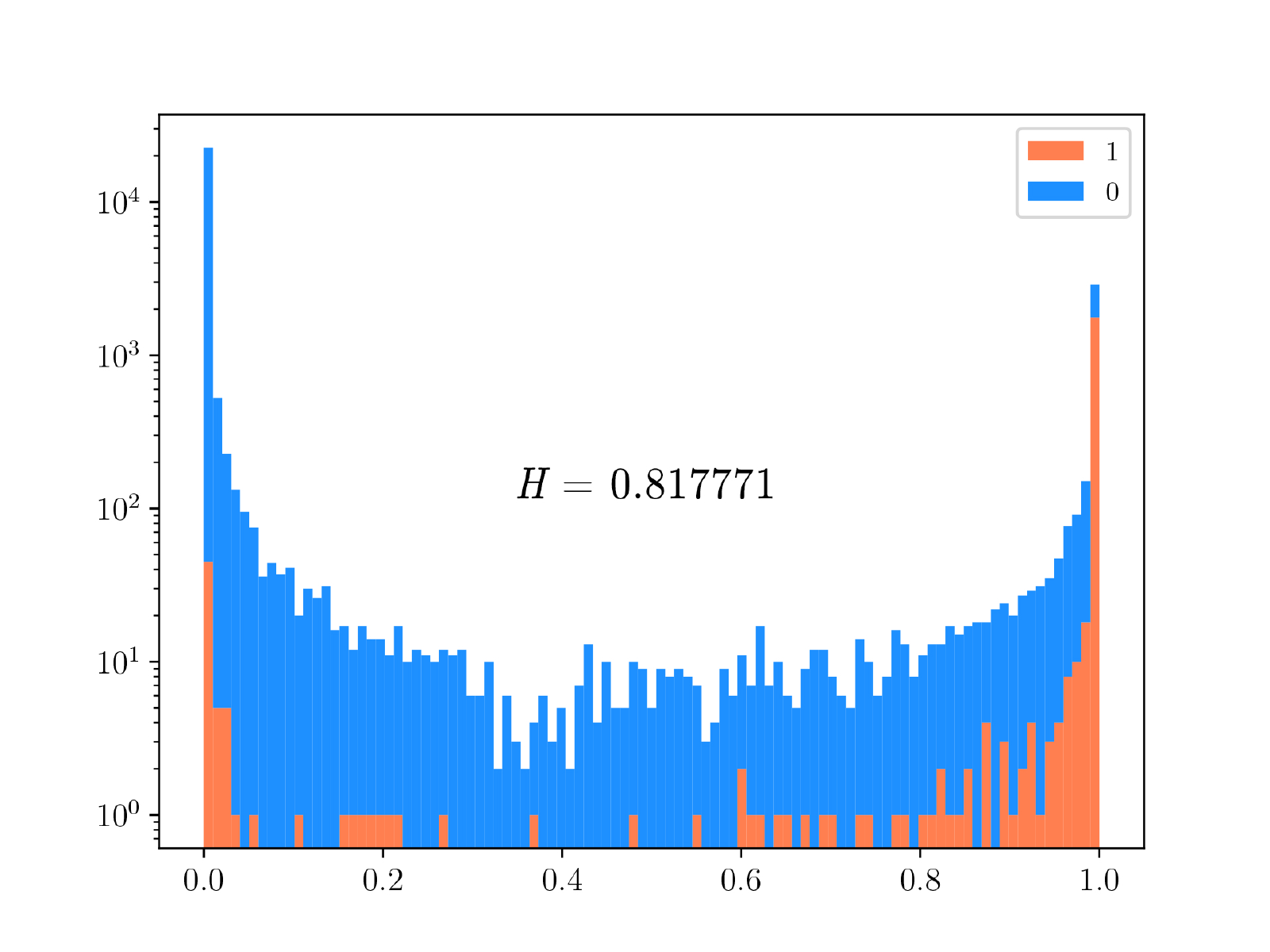} & 
\includegraphics[width=\linewidth, trim={0.5cm 0.5cm 0.5cm 0.5cm}, clip] {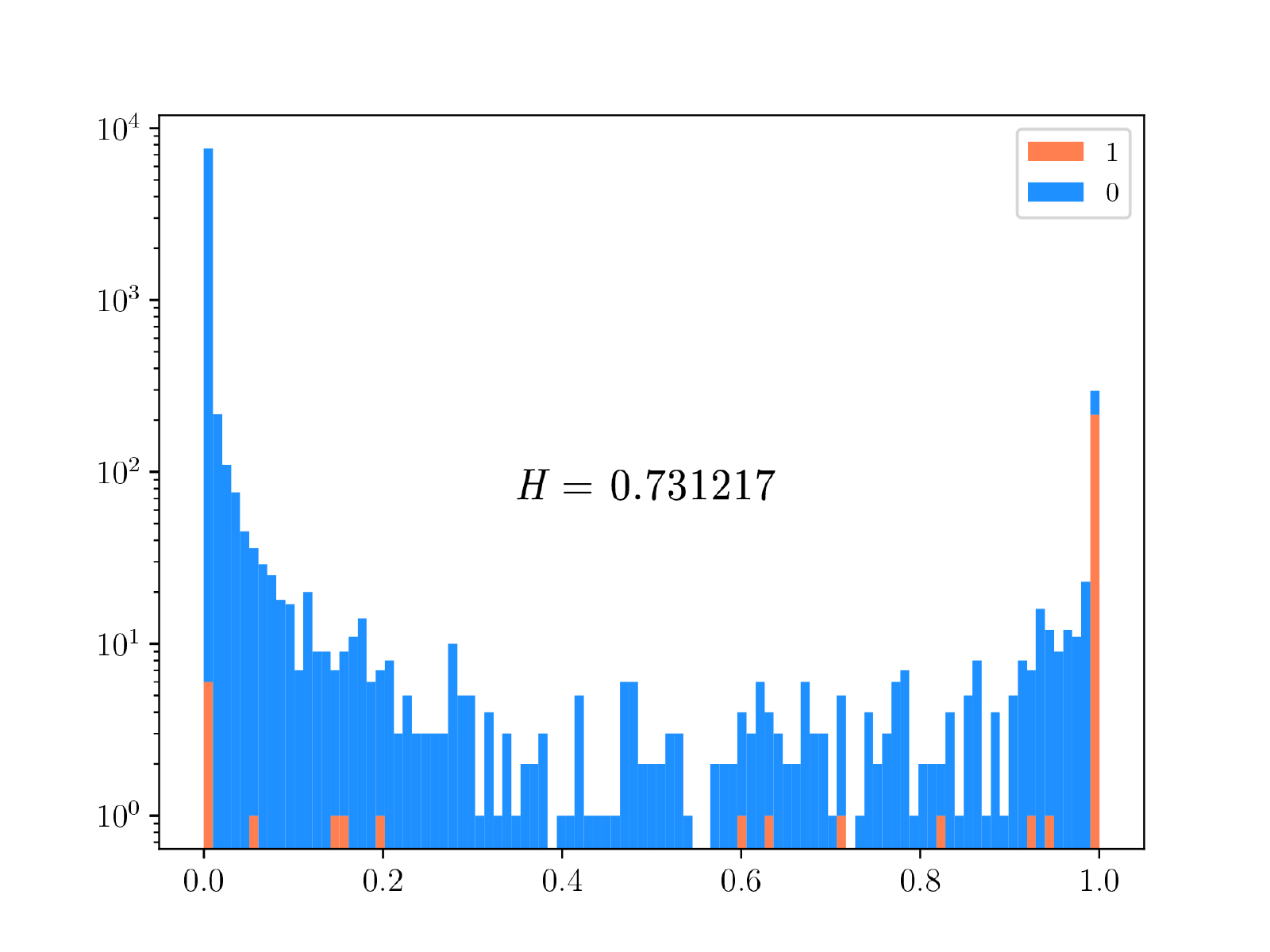} \\
\end{tabular}
\caption{Log-scaled edge score histograms displaying absolute frequencies over respective edge scores including the normalized entropy of the corresponding normalized probability distribution: Low confidence (left), high confidence (right). Orange bars represent edges with GT label 1, blue bars represent inactive edges with GT label 0.} 
\label{fig:confidence}
\end{figure*}

\begin{figure*}
\centering
\footnotesize
\begin{tabular}{P{8.5cm}P{8.5cm}}
\normalsize \ul{Low-confidence scene ($H = 0.817771$)} & \normalsize \ul{High-confidence scene ($H = 0.731216$)} \\[0.6cm]
\includegraphics[width=\linewidth, trim={5cm 8cm 5cm 8cm}, clip]{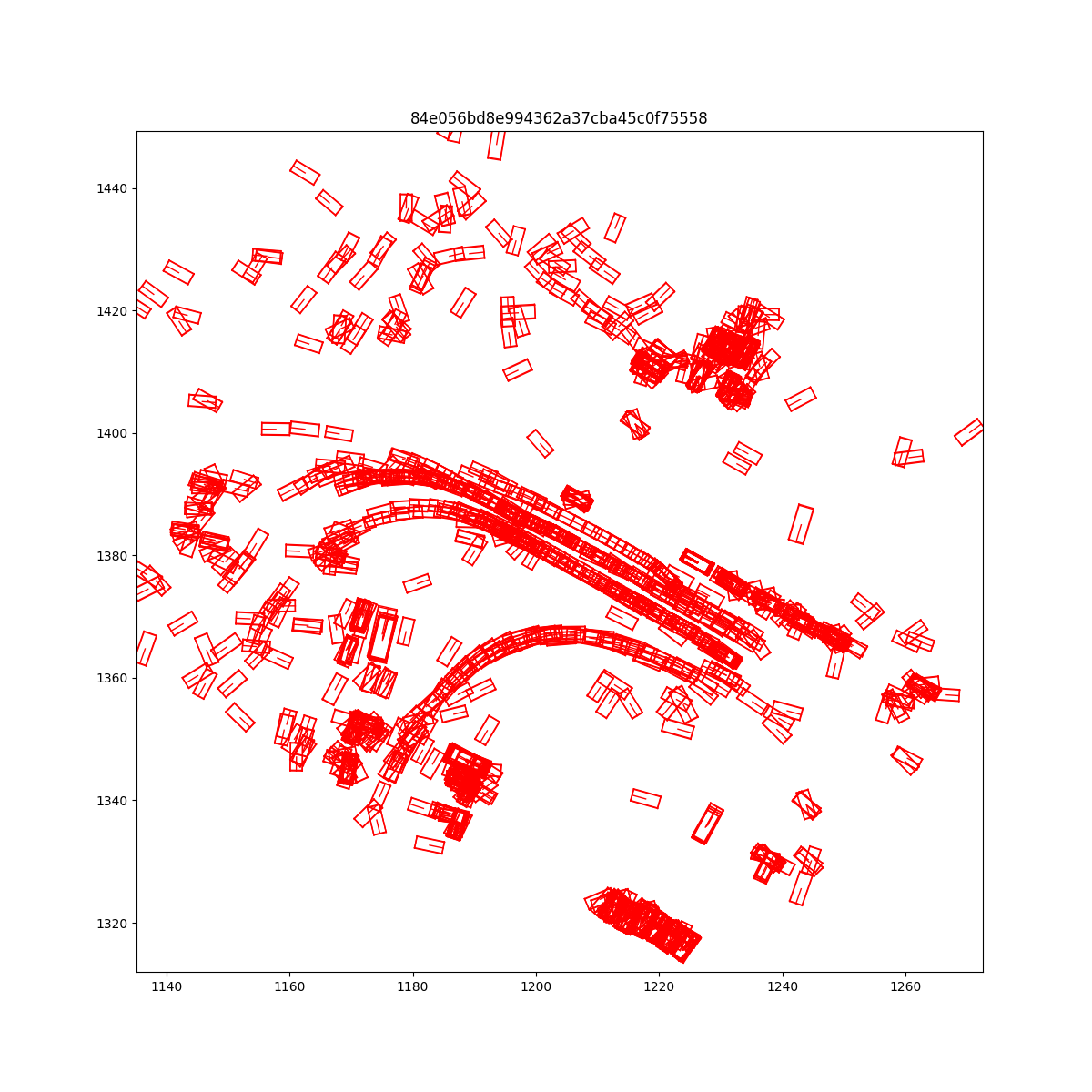} & \includegraphics[width=\linewidth, trim={5cm 8cm 5cm 8cm}, clip]{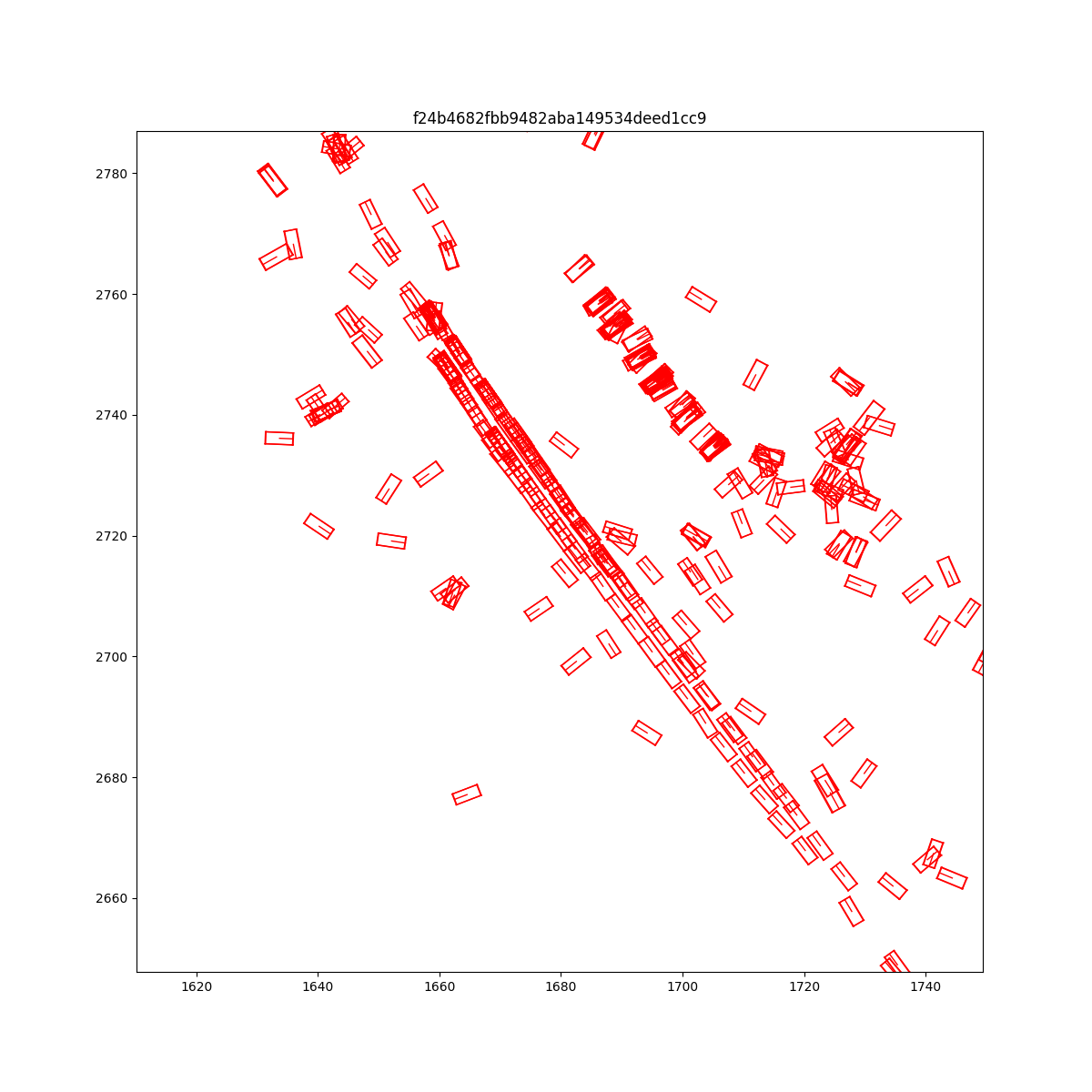} \\[0.2cm]
\multicolumn{2}{c}{a) Accumulated 3D object proposals of CenterPoint \cite{centerpoint} across 40 frames before matching.}\\[0.2cm]
\multicolumn{2}{c}{}\\
\includegraphics[width=\linewidth, trim={5cm 8cm 5cm 8cm}, clip]{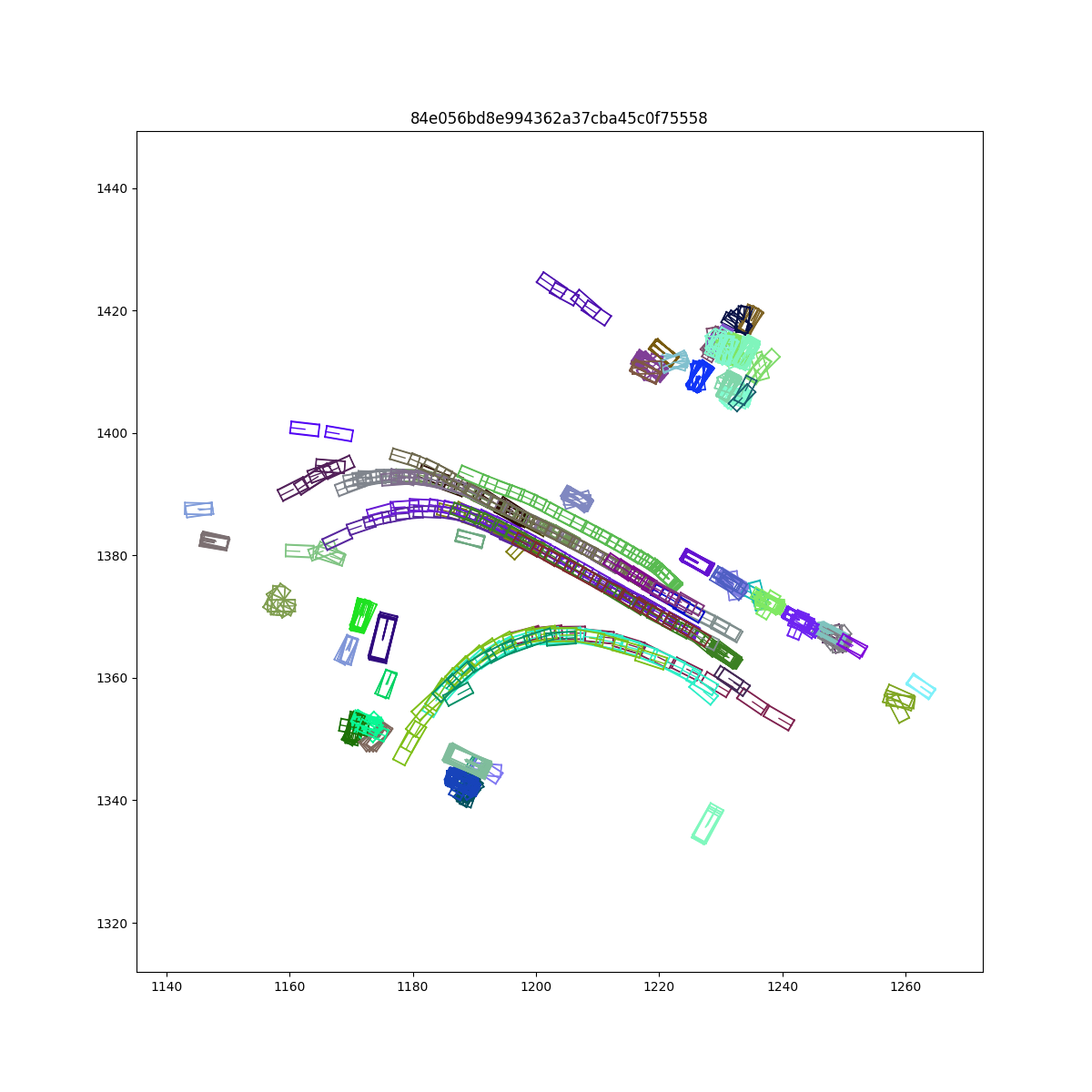} & \includegraphics[width=\linewidth,trim={5cm 8cm 5cm 8cm}, clip]{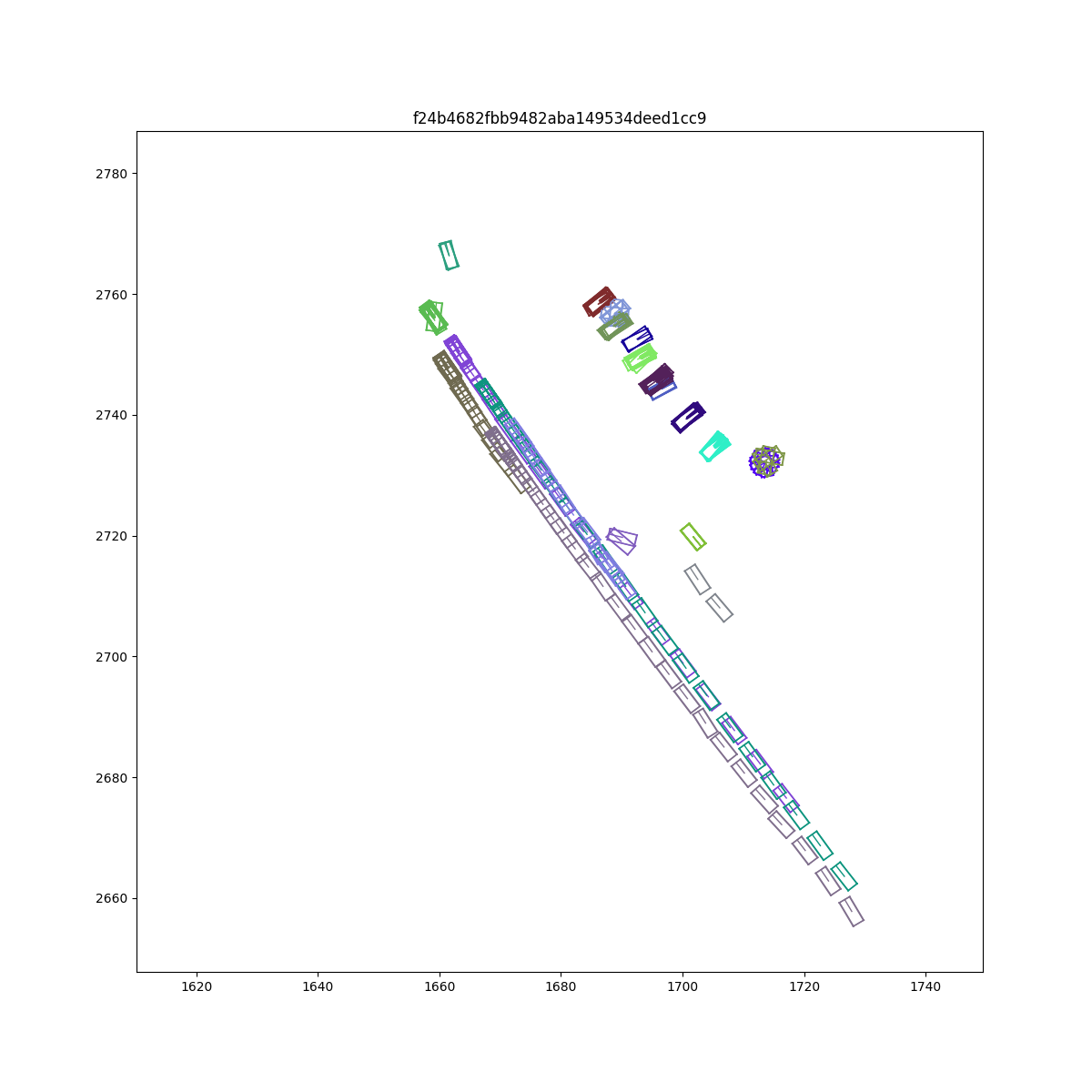} \\[0.3cm]
\multicolumn{2}{c}{b) Accumulated, non-interpolated trajectory predictions of Batch3DMOT-3D-PM-CL across 40 frames (\textit{Car} category) for two levels of confidence.} \\
\multicolumn{2}{c}{Each color denotes a particular tracking ID. Low confidence (left) and high  confidence (right).}\\[0.2cm]
\multicolumn{2}{c}{}\\
\includegraphics[width=\linewidth, trim={5cm 8cm 5cm 8cm}, clip]{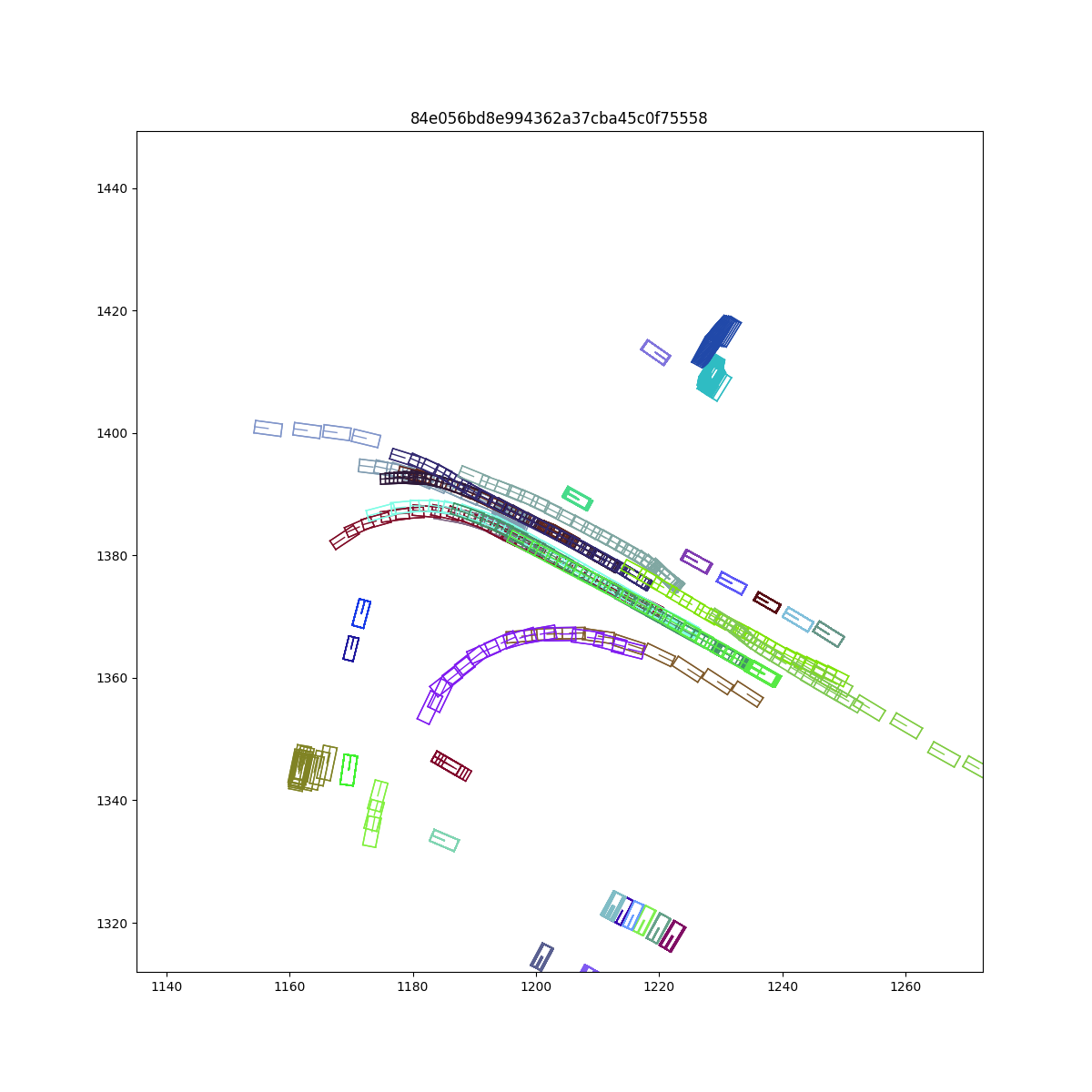} & \includegraphics[width=\linewidth, trim={5cm 8cm 5cm 8cm}, clip]{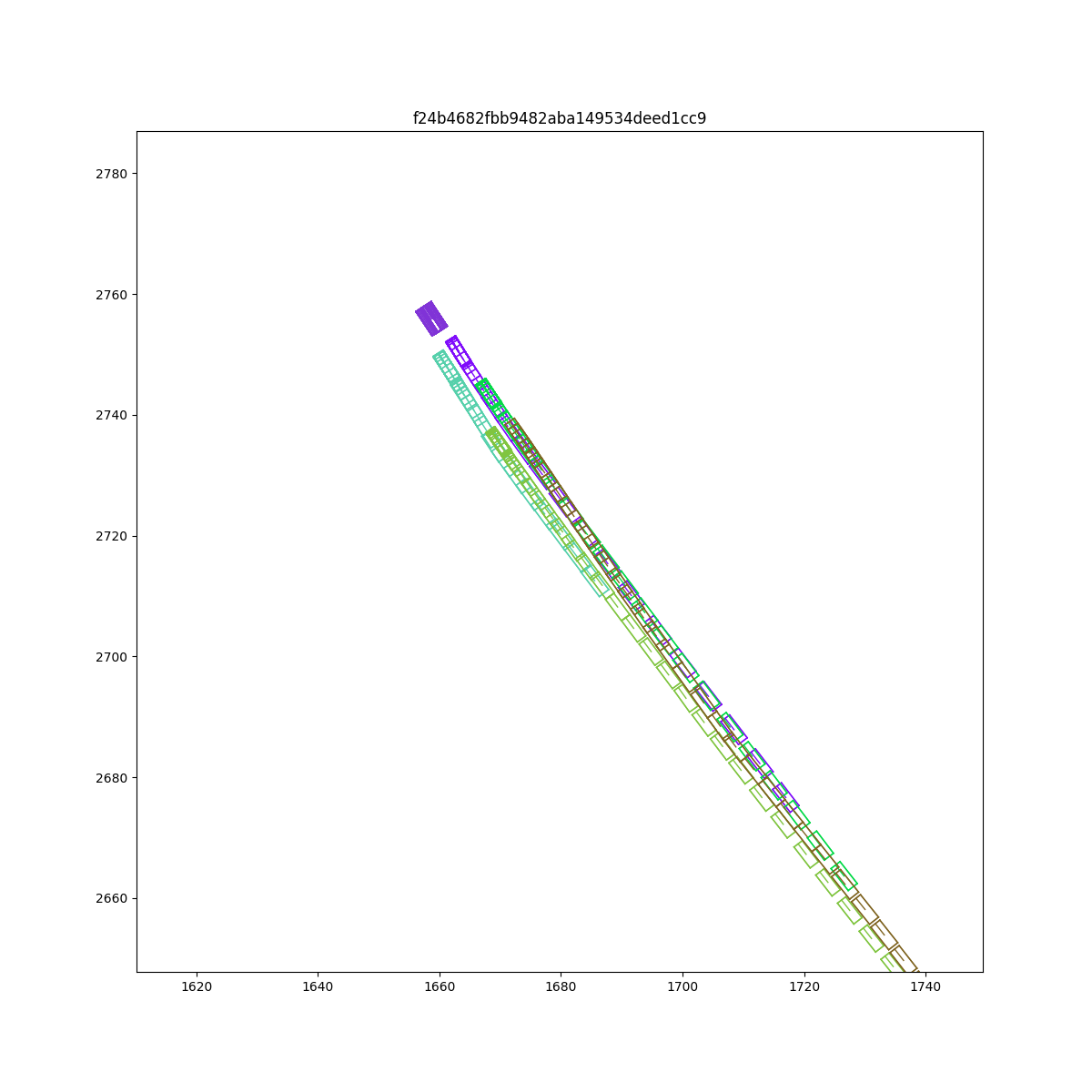} \\[0.3cm]
\multicolumn{2}{c}{c) Accumulated ground truth trajectory predictions across 40 frames (\textit{Car} category). Each color denotes a particular tracking ID.} \\[0.2cm]
\end{tabular}
\caption{Qualitative tracking results for two exemplary low confidence (left) and high confidence (right) scenes from the nuScenes validation split: Input 3D object proposals (a), unrefined predicted trajectories (b) and ground truth trajectories (c).} 
\label{fig:suppl_qual}
\end{figure*}

We propose a method to filter high confidence predictions from the test-split. Leveraging high-confidence model predictions allow to capitalize on strengths of the model instead of its weaknesses. While we do not know whether a scene itself is more or less complex, we can analyze whether the model was confident about its predictions. A batch of 5 frames yields a set of edge scores, each normalized in $\left[0,1\right]$. As each edge prediction does not represent a statement in comparison to another randomly chosen edge, each edge stands for itself. Thus, the problem of edge score prediction essentially boils down to a case of binary classification. We record the predicted edge scores per batch, construct a histogram (see \figref{fig:confidence}(a)) and normalize scores in order to construct a synthetic probability distribution. Computing the normalized entropy of the distribution as
\begin{equation}
    H = -\sum_{(j,i) \in E}{\frac{z_{(j,i)}\log z_{(j,i)}}{\log |E| }},
\end{equation}
provides a measure of tracking uncertainty. A uniform edge score distribution leads to an entropy of $H = 1$ and non-uniform distributions lead to $H < 1$. In terms of tracking, we observe that confident model predictions generally show smaller entropies (their distributions are less uniform) and vice versa (see \figref{fig:suppl_qual}). Note that the histograms depicted in \figref{fig:confidence} are log-scaled. By computing an average scene entropy using the respective batch entropies, the set of unlabeled scenes can be categorized into relatively certain and uncertain predictions. The low-confidence (high entropy) model prediction given in \figref{fig:confidence} shows a large number of false positive predictions (blue-colored edge scores in $\left[0.6,1.0\right]$), which is reflected in the cluttered tracking result depicted in \figref{fig:suppl_qual} (left). On the contrary, the high-confidence tracking prediction produces a less-cluttered set of trajectores as given in \figref{fig:suppl_qual} (right). This is further detailed in Sec.~\ref{sec:qualitative_insights_suppl}.

As means to demonstrate our findings, we report the performance of the 3D-PM model when adding either unfiltered (\textit{nusc-train + pseudo-test}) or entropy-filtered data (\textit{nusc-train + pseudo-test-entropy}) to the human-annotated training set. The entropy-filtered data contains only scenes that show a scene-entropy higher than the mean scene entropy. As presented in \tableref{tab:pseudo_training} the entropy-filtering induces a 0.2\% improvement compared to the unfiltered case. Additionally, we also trained the same 3D-PM architecture using only trajectory labels that originate from the 3D-PM-CL instance (\textit{pseudo-train + pseudo-test}), which produced a similar outcome as the human-annotations case (\textit{nusc-train}). Most notably, we do not observe any performance decline as an effect of adding weaker annotations.

We show that we can even train the 3D-PM model architecture using only pseudo-labels from the test split which contains 150 scenes. We observe an AMOTA of 0.705 (\textit{pseudo-test}) and a recall of 0.722 (see \tableref{tab:pseudo_training}). Therefore, the model shows a slight decrease in performance of about half a percent, however, using only 20\% of the data samples that are weaker than human-annotations. 

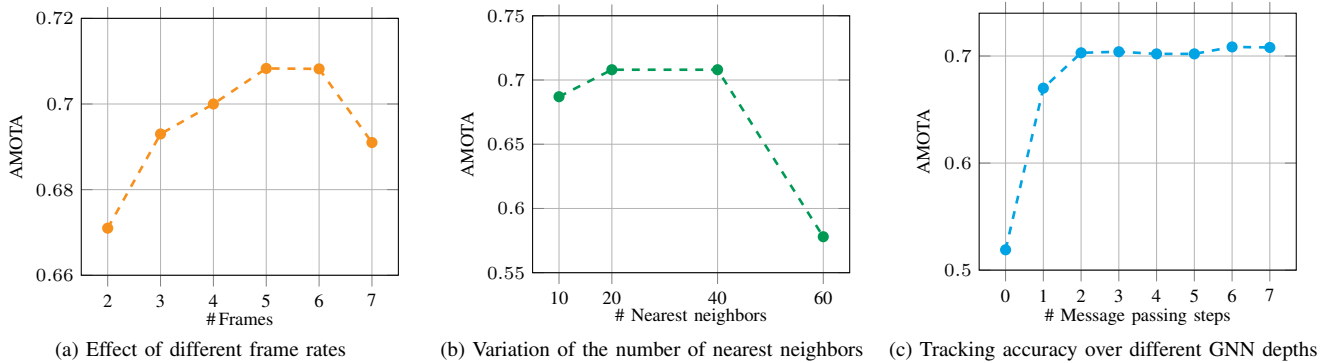
\begin{figure*}
\centering
\footnotesize
\setlength{\tabcolsep}{0.1cm}
\begin{tabular}{P{5.8cm}P{5.8cm}P{5.8cm}}
\begin{tikzpicture}
    \begin{axis}[
    	xlabel= \#\,Frames,
    	ylabel=AMOTA,
    	label style={font=\scriptsize},
    	xtick={2, 3, 4, 5, 6, 7},
        xticklabels={2, 3, 4, 5, 6, 7},
        xtick align=outside,
        xticklabel style={font=\scriptsize},
        yticklabel style={font=\scriptsize},
        ymin=0.66, ymax=0.72,
        y label style={at={(axis description cs:0.13,.5)}, anchor=south},
        x label style={at={(axis description cs:0.5,-.07)}, anchor=south},
    	grid=both,
        grid style={line width=.1pt, draw=gray!60},
    	height=5cm,
        ]
    
     \addplot[mark=*, dashed, line width=1.0pt, mark options={solid, scale=0.8}, color=BurntOrange] coordinates {
        (2, 0.671) 
    	(3, 0.693) 
    	(4, 0.700) 
    	(5, 0.7083) 
    	(6, 0.7082) 
    	(7, 0.691) 
    };
    \end{axis}
    \end{tikzpicture}
& 
\begin{tikzpicture} 
    \begin{axis}[
    	xlabel=\# Nearest neighbors,
    	ylabel=AMOTA,
    	label style={font=\scriptsize},
    	xtick={10, 20, 40, 60},
        xticklabels={10, 20, 40, 60},
        xticklabel style={font=\scriptsize},
        yticklabel style={font=\scriptsize},
        xtick align=outside,
        ymin=0.55, ymax=0.75,
        y label style={at={(axis description cs:0.13,.5)}, anchor=south},
        x label style={at={(axis description cs:0.5,-.08)}, anchor=south},
    	grid=both,
        grid style={line width=.1pt, draw=gray!60},
    	height=5cm,]
    \addplot[mark=*, dashed, line width=1.0pt, mark options={solid, scale=0.8}, color=ForestGreen] coordinates {
        (10, 0.687) 
        (20, 0.708) 
        (40, 0.708) 
    	(60, 0.578) 
    };
    \end{axis}
    \end{tikzpicture}
& 
\begin{tikzpicture} 
    \begin{axis}[
    	xlabel=\# Message passing steps,
    	ylabel=AMOTA,
    	xlabel style={font=\scriptsize},
    	ylabel style={font=\scriptsize},
    	xtick={0, 1, 2, 3, 4, 5, 6, 7},
        xticklabels={0, 1, 2, 3, 4, 5, 6, 7},
        xtick align=outside,
        xticklabel style={font=\scriptsize},
        yticklabel style={font=\scriptsize},
        ymin=0.5, ymax=0.74,
        y label style={at={(axis description cs:0.16,.5)}, anchor=south},
        x label style={at={(axis description cs:0.5,-.09)}, anchor=south},
    	grid=both,
        grid style={line width=.1pt, draw=gray!60},
    	height=5cm,]
    \addplot[mark=*, dashed, line width=1.0pt, mark options={solid, scale=0.8}, color=Cerulean] coordinates {
        (0, 0.519) 
        (1, 0.670) 
        (2, 0.703) 
    	(3, 0.704) 
    	(4, 0.702) 
    	(5, 0.702) 
    	(6, 0.7085) 
    	(7, 0.708) 
    };
    \end{axis}
    \end{tikzpicture} \\
(a) Effect of different frame rates & (b) Variation of the number of nearest neighbors & (c) Tracking accuracy over different GNN depths \\
\end{tabular}
\caption{Additional ablation study on the main Batch3DMOT model parameters. All chosen parameters stay constant apart from the one varied while its effect is measured using the AMOTA tracking score on the nuScenes validation set. The investigated model is the Batch3DMOT-3D-PM variant.} 
\label{fig:param_study}
\vspace*{0.2cm}
\end{figure*}

\subsection{Training an Online Kalman Filter Tracking Model}
\label{sec:pseudo_prob3dmot}
In addition to the pseudo-label training scheme, the experiments presented in this section use Kalman filter covariance matrix estimation introduced for Prob3DMOT~\cite{chiu2021probabilistic} using the model generated pseudo-labels. The postprocessing steps outlined in \secref{sec:postprocessing} are adopted in the same manner as in the experiment described above. We use a conjunction of pseudo-labeled nuScenes test data and human-annotations on the nuScenes train set to estimate the state uncertainty covariance $\Sigma$, the observation noise covariance $R$, and process uncertainty covariance $Q$ used by Prob3DMOT~\cite{chiu2021probabilistic}.

The results are presented in \tableref{tab:pseudo_prob3dmot}. We observe a notable 1\% gain in overall tracking accuracy compared to the standard case when using both the original training set and the pseudo-labels (\textit{nusc-train + pseudo-test}). Especially the \textit{Bicycle, Motorcycle} and \textit{Truck} classes show performance improvements. We observe only a small overall decrease (-1.3\%) when using pseudo-labels generated for both training (\textit{pseudo-train}) and test set (\textit{pseudo-test}). This demonstrates the efficacy of using pseudo-labels for \textit{training} Kalman filters, based on data statistics. Analogous to the previous experiment, we do not observe significant performance decreases due to weaker annotations.

\section{Additional Ablation Study}
\label{sec:additional_ablation}
In this section, we present an additional ablation study on the main hyperparameters of the proposed model. The most influential parameters in the Batch3DMOT framework are the number of frames, the number of nearest neighbors, and the GNN depth (the number of message passing steps). In order to identify suitable parameters (see Sec. \ref{sec:experiments}), we perform a parameter study on these variables. All experiments originate from model trainings on the nuScenes train set and evaluated on the validation split. In each study, we only vary the parameter being ablated and keep all the other hyperparameters fixed. 

\subsection{Number of Frames}
\label{subsec:abl_frames}
The number of frames per batch determine whether the FN detections based on occlusions or FP detections due to, e.g., noisy readings or misjudgement, can be recovered from. Empirically, we find that a number of 5 frames is sufficient for stable tracking (especially in case of the 2Hz framerate used in nuScenes) and still provide the grounds for comparison against 3D Kalman filter tracking models. Note that the Batch3DMOT framework only performs linear one-step interpolation of output trajectories to arrive at the results. Thus, the model itself is not capable of overcoming occlusions based on intermediate prediction-update steps as used in Kalman filtering settings. Analogously, the chosen frame rate should allow stable offline tracking with the exception of long-term occlusion handling. The study presented in \figref{fig:param_study}~(c) shows a gradual increases in tracking accuracy (measured in terms of AMOTA) until a number of 5 and 6 frames is reached under 40 nearest neighbors. For 7 frames, we observe a stark decline, presumably due to the overall number of edges rising above the critical threshold as discussed in \secref{sec:experiments}.

\subsection{Number of Nearest Neighbors}

We investigate a variation of the number of nearest neighbors (NN) leading to a edges connected in the graph construction stage. This analysis only concerns the case of semantic category-disjoint edges. Based on the findings presented in \figref{fig:param_study}~(b), an increase in the number of neighbors higher than 40 generally leads to a performance decrease. As outlined in \secref{sec:experiments}, we observe a maximum number of edges that guarantees learning success, which is exceeded in this case. On the contrary, we do not observe a performance decrease when only connecting the 20 NN over 5 frames using the proposed kinematic similarity metric (\eqref{eq:kinematic_sim_metric}), which effectively shows the efficacy of the metric. In our case, we choose 40 NN so as to overcome potential $\pm\pi$ orientation flips and velocity misjudgements stemming of noisy 3D object proposals. For 10 NN, we observe a 2.1\% performance decline compared to 20 and 40 NN (AMOTA 0.708). In general, using 20 NN over 6 frames allows further performance improvement.

\subsection{Number of Message Passing Steps}

The GNN depth is a crucial parameter determining the degree of information exchange across the proposed tracking graph. As depicted in \figref{fig:param_study}~(c), executing at least one message passing step increases tracking accuracy from AMOTA 0.519 to 0.670. Further incremental increases lead to slight improvements of the tracking performance with a maximum at 6 message passing steps (AMOTA 0.708). Compared to the other two parameters, the GNN depth bears less potential for further optimization.

\section{Qualitative Insights}
\label{sec:qualitative_insights_suppl}
The low-confidence and high-confidence scenes shown in \figref{fig:confidence} are also depicted in an accumulated BEV manner over 40 frames in \figref{fig:suppl_qual}. The left column illustrates the detections, predicted trajectories and ground truth trajectories of the low-confidence scene shown in \figref{fig:confidence} (left). The right column of \figref{fig:suppl_qual} depicts the results for a high-confidence scene.

In accordance with the edge score histogram (\figref{fig:confidence} right), we observe a much less cluttered tracking result in (\figref{fig:suppl_qual}~(b) right). On the contrary, the low confidence scene (\figref{fig:suppl_qual} left) exhibits a higher number of (presumably) false positive detections contained in the predicted trajectories. Therefore, we attribute the significant portion of incorrectly predicted edge scores with ground truth label of 0 in the range $\left[0.6, 1.0 \right]$ (see \figref{fig:confidence} left).

These findings provide a qualitative understanding of the efficacy of the introduced entropy-filtering system. Due to the unavailability of ground truth edge labels in the test split, it is infeasible to separately assess the prediction quality of either active (orange regime) or inactive (blue regime) edges in a segregated manner as shown in \figref{fig:confidence}.


\end{document}